\documentclass{article}

\usepackage[preprint]{neurips_2026}

\usepackage[utf8]{inputenc} 
\usepackage[T1]{fontenc}    
\usepackage{hyperref}       
\usepackage{url}            
\usepackage{booktabs}       
\usepackage{amsfonts}       
\usepackage{nicefrac}       
\usepackage{microtype}      
\usepackage{xcolor}         
\usepackage{graphicx}
\usepackage{wrapfig}
\usepackage{subcaption}
\usepackage{amsmath}
\usepackage{amssymb}
\usepackage{mathtools}
\usepackage{float}
\usepackage{placeins}
\usepackage[version=4]{mhchem}
\usepackage{csquotes}
\usepackage{comment}

\title{Contrastive Image-Metadata Pre-Training for Materials Transmission Electron Microscopy}

\author{%
  Georgia Channing\thanks{Equal contribution.}\,\,\thanks{Hugging Face.}\,\,\thanks{University of Oxford.} \\
  \And
  Debora Keller\thanks{Empa D\"ubendorf.} \\
  \And
  Marta D.~Rossell\footnotemark[4] \\
  \AND
  Philip Torr\footnotemark[3] \\
  \And
  Stig Helveg\thanks{Technical University of Denmark.} \\
  \And
  Henrik Eliasson\footnotemark[1]\,\,\footnotemark[5] \\
}

\begin{document}

\maketitle

\begin{abstract}
The transmission electron microscope facilitates the highest-resolution imaging of any instrument ever created, and its limiting factor is no longer spatial resolution but dose efficiency. Low electron doses avoid sample damage but produce noisy images for which, unlike in classical computer vision, there is no ground truth. Autonomous materials experimentation poses a related problem, since closed-loop instruments need representations grounded in the microscope state at acquisition. Both demand representations grounded in how an image was acquired.
We release 7,330 paired high-angle annular dark-field scanning-TEM (HAADF-STEM) images and their seven-dimensional acquisition metadata, and propose Contrastive Image-Metadata Pre-training (CIMP), a CLIP-style encoder that aligns the two modalities and reaches 84.4\% Top-1 cross-modal retrieval on a held-out split. All seven parameters are individually recoverable from the frozen visual embedding through a linear probe, and we use the embedding to condition a metadata-conditioned style-transfer model that re-renders experimental images under different acquisition parameters. Virtually scaling dwell time and beam current of low-dose images turns this model into a physics-informed denoiser; in a blind user study, experimental microscopists prefer it over the current state-of-the-art denoiser for STEM imagery on $70.2\%$ of trials.
\end{abstract}

\section{Introduction}
\label{sec:introduction}

The transmission electron microscope facilitates the highest resolution imaging of any instrument ever created, with resolutions recorded of sub-$0.5$ ångströms (or $5\times10^{-11}$ meters)~\citep{PhysRevLett.102.096101,sawada2015si}, sufficient to resolve individual atomic columns in most crystal lattices~\citep{krivanek2015aberration}.
Hence, the limiting factor of TEM is no longer spatial resolution, but instead dose efficiency, or minimising the number of electrons that hit the sample while retaining an interpretable signal. Limiting the electron dose is crucial to avoid sample damage and ensure physical relevance of observed phenomena, but naturally leads to very noisy images~\citep{EGERTON2025103819,Reimer1984}. Unlike in classical computer vision, there is no ground truth for these images, since acquiring a clean reference would itself degrade the sample. While there is optimism that machine learning workflows could mitigate the difficulties associated with low-dose imaging, a lack of open-source data means that current methods are typically trained from scratch for a specific noise level on a single or narrow set of samples. Beyond denoising specifically, the broader trajectory of materials-science instrumentation is towards autonomous scientific experimentation, in which microscopes operate in closed loops that collect, interpret, and respond to their own measurements with reduced human supervision~\citep{abolhasani2026autonomous}. The TEM's complexity, with many physically meaningful parameters tuned per acquisition, means that any such autonomous system needs a representation of each image grounded in the instrument state under which it was acquired. The same kind of representation also enhances interactive use, supplying operators with real-time tools that reason over the microscope's state during an experiment. In both modes, the missing piece is an encoder that aligns visual content with the physical parameters of the instrument.

Yet, for every published image, a massive volume of data, more than $90\%$ by some estimates~\citep{chemistry7050160}, remains unused, not because the quality of the data is bad but because TEM experiments require exploration and tuning to identify and record the golden data that eventually makes publication. Over time, TEM labs accumulate large volumes of leftover data from different users, materials samples, operating modes, and acquisition settings, and this data is rich in detail and variation, often paired with automatically saved metadata that describe the state of the microscope at acquisition. Unfortunately, this data is almost never shared as the culture of data sharing is not established in the electron microscopy community~\citep{chemistry7050160}.

\begin{wrapfigure}{r}{0.42\linewidth}
    \centering
    \includegraphics[width=\linewidth]{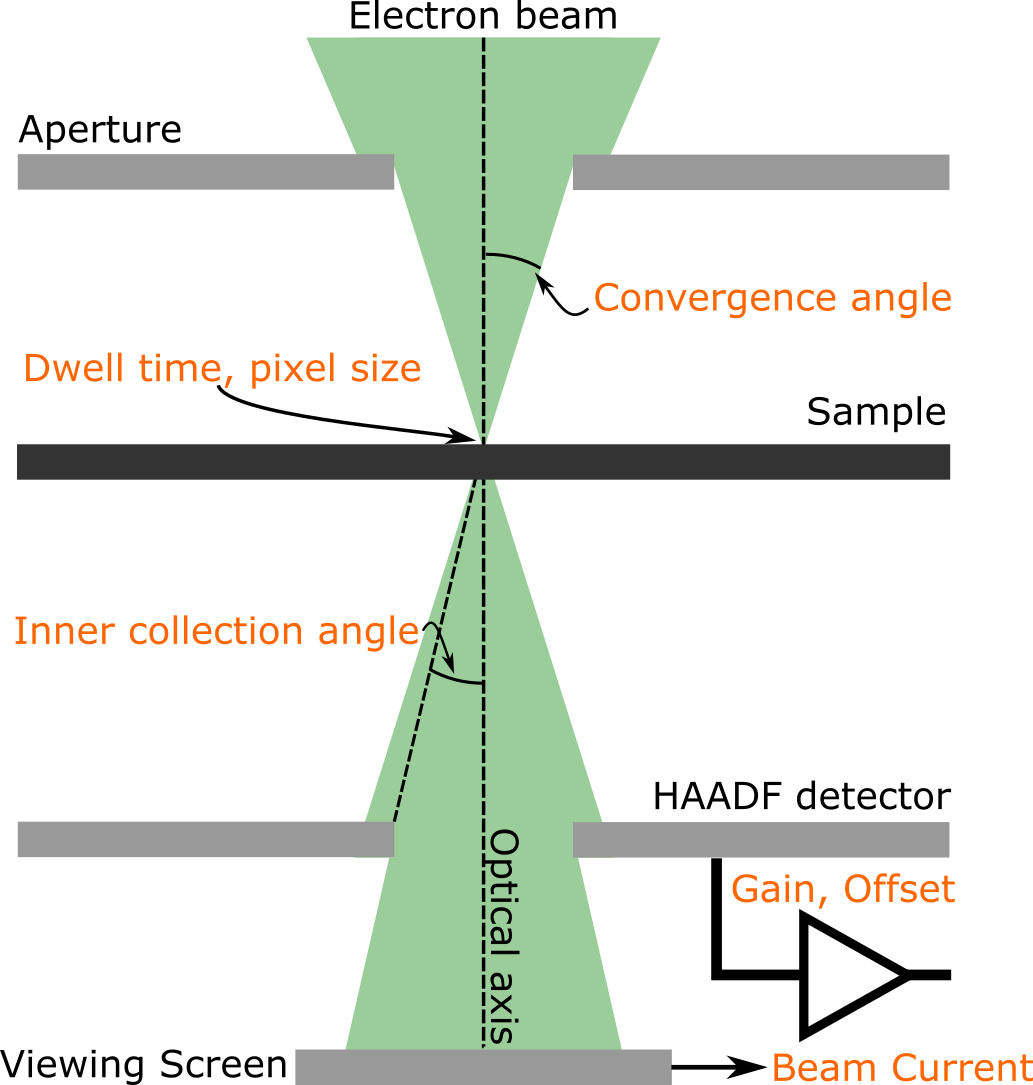}
    \caption{Diagram of a STEM with our acquisition parameters shown in orange text. Description of each acquisition parameter is provided in Appendix~\ref{app:metadata_explanation}.}
    \label{fig:stem_diagram}
\end{wrapfigure}

In this work, we address these problems through four contributions. First, we release a paired image-metadata dataset of 7{,}330 high-angle annular dark-field scanning-TEM (HAADF-STEM) electron microscopy images, each associated with seven acquisition-metadata parameters: pixel size, dwell time, beam convergence angle, beam current, detector gain, detector offset, and inner collection angle (Figure~\ref{fig:stem_diagram}). Second, we propose Contrastive Image-Metadata Pre-training (CIMP), a CLIP-style method~\citep{pmlr-v139-radford21a} that learns a joint embedding space over HAADF-STEM images and their acquisition metadata, reaching Top-1 cross-modal retrieval of $84.4\%$ on a held-out validation split, and we treat the resulting embeddings as a foundation layer for a variety of potential TEM-ML workflows rather than as infrastructure for any single downstream task. Third, linear probing shows that all seven acquisition parameters are individually recoverable from the frozen $128$-d visual embedding through Ridge regression despite the contrastive objective never supervising any metadata value directly, and we use the embedding as both a perceptual loss and a conditioning signal for a generative metadata-conditioned style-transfer model that translates an experimental image into the style it would have had under different acquisition parameters. Finally, we use the style-transfer model as a physics-informed denoiser by virtually scaling dwell time and beam current of low-dose images. In a blind user study, experimental microscopists prefer this denoiser over the state-of-the-art Noise2Void baseline on $70.2\%$ of trials.

\section{Related Work}

\paragraph{Open-Source STEM Data}
While petabytes of HAADF-STEM images are collected each year~\citep{chemistry7050160}, very few are ever open-sourced. Microscopy publications typically focus on findings from the highest-quality images, and the culture of data sharing is not well-established in the field. Some larger experimental and synthetic datasets have nevertheless been released. The closest experimental precedent is from \citet{sytwu2023segmented}, who released 407 HRTEM images of \ce{Au}, \ce{Ag}, and \ce{CdSe} nanoparticles paired with acquisition metadata including magnification and electron dose. This release was much smaller in scale and in HRTEM rather than HAADF-STEM.
Other large-scale open releases are simulated. Atomagined~\citep{schwenker2020atomagined} provides PRISM-generated HAADF-STEM images post-processed to emulate noise and distortion, covering a wide range of ICSD structure prototypes. \citet{eliasson2025multislice, eliasson2025synthetic, Eliasson3D} release a 250,000-image multislice-simulated HAADF-STEM corpus of \ce{Pt} nanoparticles for size and morphology prediction. TEMImageNet~\citep{lin2021temimagenet} offers a simulated training library for atom-resolution STEM tasks and shows that simulation-trained models can transfer to experimental data. None of these match the scale of pairing between experimental HAADF-STEM images and rich instrument metadata that we target. Existing experimental datasets tend to be small, narrow in material coverage, and rarely include the acquisition metadata needed to train models that generalise across instrument settings.

\paragraph{Embedding Models}
CIMP adapts the CLIP image--text pretraining framework~\citep{pmlr-v139-radford21a} to pairs of HAADF-STEM images and their acquisition metadata. The closest structural analogue in any microscopy domain is CLOOME~\citep{sanchezfernandez2023cloome}, which contrastively aligns fluorescence cell-painting images with chemical-structure embeddings and demonstrates cross-modal retrieval across a large bioimaging database. ContIG~\citep{taleb2022contig} is the closest conceptual parallel: a medical-imaging model that contrastively aligns retinal images with a structured physical-annotation vector (SNP arrays, polygenic risk scores). To our knowledge, however, no prior work applies this or any similar framework to electron microscopy (EM).

\paragraph{Image-to-Image Translation in Electron Microscopy}
Prior work has applied image-to-image networks to EM data in two main directions: to bridge the simulation-to-experiment gap so that synthetic training data gains realism, and for denoising of experimental acquisitions. To close the simulation-to-experiment gap, \citet{Khan2023} use a CycleGAN~\citep{zhu2017cyclegan} with a reciprocal-space discriminator to add realistic high-frequency content to simulated STEM images so that a downstream segmentation network can transfer to real experiments. \citet{EliassonCycleGAN} use a CycleGAN to translate between simulated and experimental HAADF-STEM of \ce{Pt} nanoparticles, showcasing both denoising and realistic noise addition. And in another work, \citet{Lobato2024} train a GAN denoiser on synthetic image pairs with intricately modeled noise. These approaches are all one-to-one and agnostic to microscope settings (i.e., a dedicated network is trained per source/target domain pair, with no natural way to parameterise the target style). Simulating a different dwell time or detector gain with any previous method requires training a new CycleGAN with a new paired or unpaired corpus at that setting. The style-transfer network we present downstream of CIMP is instead one-to-many: a single generator is conditioned on continuous acquisition metadata through CIMP embeddings such that any target instrument setting can be queried at inference without retraining.

\paragraph{STEM Denoisers}
Microscopists are always pushing the limit of their detector in order to minimize the electron dose. With those ever diminishing electron doses, the need for appropriate denoisers to make low-dose data interpretable and easy to use, is constant. Self-supervised denoisers in the Noise2Noise / Noise2Void / UDVD family~\citep{pmlr-v80-lehtinen18a,krull2019noise2void,Sheth_2021_ICCV} have been popular approaches to denoising STEM imagery, with several recent applications to atomic-resolution data~\citep{park_denoise,Thornley2026}. However, these models are heavily specialised for single noise profiles and materials samples, some of which (such as Noise2Noise) further require time-series data where noise changes but image content does not (i.e., true image pairs). Noise2Void-based techniques that can work on single or unpaired images instead introduce checkerboard artefacts when trained to convergence and lose image detail if pixel size is not significantly smaller than the atomic features the experimentalist is interested in. GAN-based denoisers have also been explored. Noise2Atom~\citep{FengNoise2Atom} is an unsupervised GAN-based denoiser trained on noisy acquisitions of sub-nm catalytic clusters, and is, to our knowledge, the only STEM denoiser to release the data with which it was trained. As far as the authors are aware, there exist no \textit{general} STEM denoisers in the sense that they are informed about the acquisition settings and pixel size related to the images and can produce good results across instrument settings.

\section{Data Contribution}

In this work, we introduce 7,330 HAADF-STEM images with paired metadata from a Titan microscope at an anonymised materials-research laboratory. The dataset covers more than 30 distinct materials systems, predominantly heterogeneous catalysts (\ce{Pt/CeO2}, \ce{Pd/ZrO2}, \ce{Ru/TiO2}, \ce{In2O3/ZrO2}, etc.), with more than 130 material samples synthesized by different techniques and with different metal loading, imaged with a variety of instrument settings. The full size images have variable sizes between $256 \times 256$ and $4096 \times 4096$, with $83\%$ at $2048 \times 2048$. We show the distribution of metadata parameters in our dataset in Figure~\ref{fig:metadata_distribution} in Appendix~\ref{app:metadata_explanation}, as well as a description of each of the metadata variables (also in Appendix~\ref{app:metadata_explanation}). We additionally solicited the creation of several \textit{evaluation sets}, which are series of true experimental images captured of the same sample in quick succession while varying a single metadata parameter. These series sweep dwell time, gain, and offset respectively. We hold this data out from training and retain it solely for qualitative comparison with our model outputs, as shown in Figure~\ref{fig:sweeps}. Outside of these sets, images were captured as the operator saw fit, so the corpus contains no paired or matched groupings, only a broad collection of samples spanning varied acquisition parameters. We will release this data (along with our models and code) with a \texttt{CC-by-4.0} license once the blind review period is concluded.

\section{Contrastive Image-Metadata Pretraining (CIMP)}

We train a CLIP-style contrastive image-metadata embedding model on our dataset of 7,330 HAADF-STEM images and their paired metadata. We evaluate the performance of our embedding model, CIMP, by measuring retrieval accuracy at \textit{Top-1}, \textit{Top-5}, and \textit{Top-10}. \textit{Top-1} denotes the percentage of queries for which the correct match is ranked first among the retrieved results. \textit{Top-5} and \textit{Top-10} indicate the percentage of queries for which the correct match appears within the top 5 and top 10 results, respectively.

For training, each HAADF-STEM image is normalised from 16-bit unsigned integers to a floating-point tensor in $[0,1]$ and a single random $256 \times 256$ crop is extracted per image per epoch, augmented with a horizontal flip (probability $0.5$) and a random multiple-of-$90^\circ$ rotation. The accompanying metadata vector contains seven acquisition parameters. The four dimensions that span several orders of magnitude (pixel size, dwell time, convergence angle, beam current) are stored in $\log_{10}$ space, and all seven dimensions are z-scored using statistics computed across the training corpus before being passed to the metadata encoder.

To train CIMP, we split our dataset into training and validation splits of size 6{,}597 and 733, respectively, with a fixed seed. The split is stratified at the full-size image level such that all crops drawn from a given source image are assigned to the same split. Thus, retrieval metrics measure generalisation to unseen acquisitions rather than memorisation of within-image content (see Appendix~\ref{app:dataset_construction} for details). We sample $1$ random crop per image per training epoch, which yields effectively unlimited unique patches across our training.

CIMP is trained with the symmetric InfoNCE objective~\citep{oord2018cpc} as used in CLIP~\citep{pmlr-v139-radford21a}. For a batch of $B$ paired samples, the visual encoder maps each image to an $L_2$-normalised embedding $f_i \in \mathbb{R}^{128}$ and the metadata encoder maps each metadata vector to an $L_2$-normalised embedding $g_j \in \mathbb{R}^{128}$. With pairwise cosine similarity $s_{ij} = f_i^\top g_j$, the loss is
\begin{equation}
\mathcal{L}_{\text{CIMP}} = -\frac{1}{2B} \sum_{i=1}^{B} \left( \log \frac{\exp(\tau s_{ii})}{\sum_{j=1}^{B} \exp(\tau s_{ij})} + \log \frac{\exp(\tau s_{ii})}{\sum_{j=1}^{B} \exp(\tau s_{ji})} \right),
\end{equation}
the average of two cross-entropies that pull each paired $(f_i, g_i)$ together while pushing all unpaired samples in the batch apart. The temperature $\tau$ is learnable and initialised at $0.1$.

To find the best architecture and hyper-parameters for CIMP, we compare four image-encoder backbones (ResNet-18~\citep{he2016resnet}, ViT~\citep{dosovitskiy2021vit} trained from scratch, ImageNet-pretrained ViT finetuned end-to-end, and ImageNet-pretrained ViT with the backbone frozen), four batch sizes ($128,\ 256,\ 512,\ 1024$), five metadata encoder sizes ($64, 128,\ 256,\ 512,\ 1024$), and four crop sizes ($128,\ 256,\ 512,\ 1024$). The results and discussion of all these studies is reported in Appendix~\ref{app:pretraining_ablations}.

We find the best performance in and adopt going forward a ResNet-18 model trained from scratch with a single-channel input with a crop size of $256\times256$, a batch size of $512$, and a $3$-layer MLP metadata encoder of width $256$. We train for 1000 epochs and achieve final Top-1 of $0.8438$, Top-5 of $0.9688$, and Top-10 of $0.9844$ on our held-out validation set. Notably, an ImageNet-pretrained ViT finetuned end-to-end on our HAADF-STEM data (Top-1 $82.8\%$) does not outperform our ResNet-18 trained from scratch (Top-1 $84.4\%$) on this task (Appendix~\ref{app:pretraining_ablations}), which conveys the need for HAADF-STEM image priors given the difference in distributions between natural images and atomic-scale images. Qualitative t-SNE projections of the frozen CIMP visual embeddings, coloured by each metadata dimension and by acquisition session, are provided in Appendix~\ref{app:cimp_embedding_visualization}.

\section{Style Transfer, Denoising, and Metadata Recovery}

We evaluate CIMP in three stages, each reported as its own subsection below. We first characterise the learned representation by asking how much of the raw acquisition metadata can be recovered from the frozen visual embedding alone through linear probing. Then, we use the metadata embeddings as a conditioning signal and perceptual loss for a metadata-conditioned style-transfer model. As discussed in Section~\ref{sec:introduction}, noise in HAADF-STEM images increases with shorter dwell time and lower beam current, often used to preserve sample integrity, as fewer electrons impinge on the detector at each raster position of the STEM scan.  To use our style-transfer model as a physics-informed denoiser, we change the ``style'' of the image to one with higher dwell time and beam current. We compare the fidelity of our style transfer model to experimentally varying the acquisition parameters with our evaluation set. We evaluate its performance as a denoiser by comparing its output quantitatively and qualitatively against the CycleGAN-based Noise2Atom denoiser from ~\citet{FengNoise2Atom} and the state-of-the-art, Noise2Void-based denoiser for HAADF-STEM images from ~\citet{Thornley2026}, and we additionally validate it through a blind user study with experimental microscopists.

\subsection{Metadata Recovery}
\label{sec:linear_probe}

For CIMP embeddings to serve as a foundation layer that downstream models can condition on, the representation must encode physically meaningful axes of microscope state, not just enough information to re-pair an image with its own metadata at retrieval time. To probe this, we extract $128$-d visual embeddings from the frozen CIMP encoder for every image in the train and validation splits (6{,}597 / 733 images, the same split as CIMP pre-training). We then fit seven independent Ridge regressors ($\alpha = 1.0$) on the training embeddings, one per metadata dimension, each predicting a single z-scored target value. We report per-dimension $R^2$ and symmetric mean absolute percentage error (SMAPE) on the held-out validation split, with SMAPE computed in physical units after inverting both the z-score and the $\log_{10}$ transform applied during preprocessing. Since Ridge has a closed-form solution, the probe measures how much metadata information is linearly present in the embedding without additional nonlinear decoding capacity.

\begin{wraptable}{r}{0.55\linewidth}
\centering
\caption{Per-dimension linear-probe recovery from the frozen CIMP visual embedding. Values are mean $\pm$ standard error across $2000$ bootstrap resamples; the Mean row reports a point estimate. Dimensions marked $^\dagger$ are stored in $\log_{10}$ space.}
\label{tab:linear_probe}
\footnotesize
\setlength{\tabcolsep}{3pt}
\renewcommand{\arraystretch}{1.15}
\begin{tabular}{lcc}
\toprule
\textbf{Dimension} & $R^2\uparrow$ & SMAPE$\downarrow$ \\
\midrule
Pixel size$^\dagger$        & $0.72 \pm 0.03$ & $41.8 \pm 1.2\%$ \\
Dwell time$^\dagger$        & $0.78 \pm 0.01$ & $28.4 \pm 0.9\%$ \\
Convergence angle$^\dagger$ & $0.53 \pm 0.04$ & $13.0 \pm 0.4\%$ \\
Beam current$^\dagger$      & $0.65 \pm 0.03$ & $34.6 \pm 1.1\%$ \\
Gain                        & $0.84 \pm 0.01$ & $5.3 \pm 0.2\%$ \\
Offset                      & $0.80 \pm 0.05$ & $9.2 \pm 1.2\%$ \\
Inner coll.\ angle          & $0.49 \pm 0.05$ & $9.5 \pm 0.4\%$ \\
\midrule
Mean                        & 0.69            & 20.3\%            \\
\bottomrule
\end{tabular}
\end{wraptable}
All seven metadata parameters are linearly decodable from the frozen $128$-d embedding (Table~\ref{tab:linear_probe}). Gain, offset, inner-collection angle, and convergence angle are recovered to within 5--13\% SMAPE; their limited dynamic range and direct effect on global image statistics keep percentage errors low. The three $\log_{10}$-transformed dimensions with wide dynamic range (pixel size, dwell time, beam current) show inflated SMAPE of 28--47\% despite $R^2 \in [0.66, 0.78]$, reflecting the multiplicative effect of re-exponentiating $\log_{10}$-space residuals.

The contrastive objective we used never supervises any metadata value directly; it only asks visual embeddings to align with the MLP projection of the metadata vector. Yet seven physically distinct acquisition parameters are independently recoverable from the frozen $128$-d embedding through a closed-form linear probe, evidence that contrastive image-metadata alignment yields a factored, physics-aligned representation as a byproduct of retrieval training. To our knowledge, this is the first demonstration of such a factored representation for electron microscopy, and it is what makes CIMP embeddings useful as a conditioning signal beyond the retrieval task itself.

\subsection{Style Transfer}
\label{sec:style_transfer}
In practice, the most useful denoiser for an experimental microscopist is one that runs while they are at the microscope, quicker than the frame rate of the acquisition, so that low-dose images can be interpreted in time to inform the next decision. This rules out methods whose inference cost is dominated by iterative refinement. We additionally cannot rely on paired source-target data, since acquiring matched low-dose and high-dose images of the same field would itself degrade the sample, and our training corpus of $\sim$7{,}000 images is small by computer-vision standards. These constraints motivate a GAN-based architecture. GANs admit one-shot inference without iterative ODE solving, train reliably on small datasets, integrate naturally with the multi-task LPIPS-based objective we use, and allow us to jointly optimise for the correct metadata style, content preservation, and image realism without true image-metadata pairs. By contrast, Palette-style image-to-image diffusion~\citep{saharia2021palette} requires paired source-target examples that we do not have, and flow matching~\citep{lipman2022flow} is natively a noise-to-image generative method rather than an image-to-image translator that can be conditioned on metadata.

Our style transfer model takes an input image $x$ along with a source embedding $e_{\text{id}}$ of the image's original metadata and a target embedding $e_{\text{tgt}}$ of the shuffled metadata we want the output to match. Both are produced by the frozen CIMP metadata encoder from raw acquisition-metadata vectors. The model then produces an image that preserves the content of $x$ but matches the style associated with $e_{\text{tgt}}$. We implement this with a U-Net~\citep{ronneberger2015unet} generator $G$ whose convolutional blocks are conditioned on the embeddings via FiLM layers~\citep{perez2018film}. This allows the metadata signal to modulate features at every scale, while the U-Net skip connections carry content information through unchanged. 

We train $G$ against four objectives. The adversarial loss $\mathcal{L}_{\text{LSGAN}}$, in the Least Squares GAN formulation~\citep{mao2017least}, drives outputs toward the distribution of real micrographs. The remaining three components are defined as:
\setlength{\abovedisplayskip}{4pt}\setlength{\belowdisplayskip}{4pt}
\begin{align}
    \mathcal{L}_{\text{cyc}} &= \text{LPIPS}\big(G(G(x, [e_{\text{tgt}}; e_{\text{id}}]), [e_{\text{id}}; e_{\text{tgt}}]),\ x\big), \\
    \mathcal{L}_{\text{id}}  &= \text{LPIPS}\big(G(x, [e_{\text{id}}; e_{\text{id}}]),\ x\big), \\
    \mathcal{L}_{\text{emb}} &= \big\| e_{\text{tgt}} - \text{CIMP}_{\text{vis}}(G(x, [e_{\text{tgt}}; e_{\text{id}}])) \big\|_2^2, \\
    \mathcal{L}_{\text{GAN\_total}} &= \mathcal{L}_{\text{LSGAN}} + \lambda_1 \mathcal{L}_{\text{cyc}} + \lambda_2 \mathcal{L}_{\text{id}} + \lambda_3 \mathcal{L}_{\text{emb}},
\end{align}
where $\mathcal{L}_{\text{cyc}}$ enforces content preservation by requiring a round trip (source to target, then target back to source) to recover $x$, following the cycle-consistency idea of \citet{zhu2017cyclegan} with LPIPS~\citep{zhang2018perceptual} as the distance metric; $\mathcal{L}_{\text{id}}$ discourages $G$ from modifying images whose target matches their source; and $\mathcal{L}_{\text{emb}}$ anchors the output to $e_{\text{tgt}}$ in the CIMP visual-embedding space, ensuring the generated image matches the target style. We set $\lambda_1 = 0.3$, $\lambda_2 = 0.1$, and $\lambda_3 = 1.0$ empirically; see Appendix~\ref{app:gan_ablations} for the loss-weight ablation that motivated these values.

\begin{table}[H]
\centering
\begin{minipage}[t]{0.49\linewidth}
\centering
\caption{Pixel-level fidelity on the three held-out evaluation series, averaged over all ordered pairs $(x_i, x_j)$ within each series. CIMP GAN rows compare $G(x_{\mathrm{src}}, e_{\mathrm{tgt}}, e_{\mathrm{id}})$ to the real reference $x_j$; Eval series rows compare the unmodified $x_i$ to $x_j$.}
\label{tab:style_transfer_pixel}
\vspace*{5pt}
\footnotesize
\setlength{\tabcolsep}{3pt}
\renewcommand{\arraystretch}{1.0}
\begin{tabular}{llrrr}
\toprule
\textbf{Series} & \textbf{Model} & SSIM$\uparrow$ & PSNR$\uparrow$ & LPIPS$\downarrow$ \\
\midrule
Dwell   & CIMP GAN          & 0.311 & 22.82 & 0.188 \\
        & Eval series  & 0.324 & 24.07 & 0.187 \\
\midrule
Gain    & CIMP GAN          & 0.118 & 11.03 & 0.322 \\
        & Eval series  & 0.100 & 10.48 & 0.369 \\
\midrule
Offset  & CIMP GAN          & 0.153 & 10.77 & 0.364 \\
        & Eval series  & 0.151 & 10.30 & 0.367 \\
\midrule
Overall & CIMP GAN          & 0.211 & 16.13 & 0.273 \\
        & Eval series  & 0.210 & 16.38 & 0.289 \\
\bottomrule
\end{tabular}
\end{minipage}\hfill
\begin{minipage}[t]{0.49\linewidth}
\centering
\caption{Kernel Inception Distance (KID) between four pairs of image sets. Subset size $21$ for the first three rows (bounded by the eval-set size) and $50$ for the noise floor; the last three rows average across $5$ disjoint training pools of $500$ images with pinned seeds.}
\label{tab:style_transfer_kid}
\vspace*{52pt}
\footnotesize
\setlength{\tabcolsep}{4pt}
\renewcommand{\arraystretch}{1.15}
\begin{tabular}{lr}
\toprule
\textbf{Comparison} & \textbf{KID}$\downarrow$ \\
\midrule
CIMP GAN vs.\ eval set                 & $-0.016 \pm 0.009$ \\
CIMP GAN vs.\ training                 & $0.185 \pm 0.009$ \\
Eval set vs.\ training (content floor) & $0.189 \pm 0.010$ \\
Training vs.\ training (noise floor)   & $0.0003 \pm 0.0004$ \\
\bottomrule
\end{tabular}
\end{minipage}
\end{table}

To evaluate CIMP GAN's performance on style transfer, we shuffle the metadata in the validation set to assign a target style for each image. Style-transferred images are then passed through the CIMP visual encoder to generate embeddings, and we calculate the cosine similarities between these and the target metadata embeddings. If the generator has learned the metadata-to-style mapping, the cosine similarities between output images and their target style should be high while the similarity with other metadata embeddings should be low. As shown in Figure~\ref{fig:sim_dist_gen_images}, the mean similarity of correct pairs exceeds that of incorrect pairs by $0.523$ on the cosine-similarity scale (bounded in $[-1, 1]$), demonstrating that the generator aligns image style with the provided conditioning embedding.

Our evaluation set contains series of real experimental images acquired at different metadata settings for dwell, gain, and offset. For every ordered pair of images $(x_i, x_j)$ within a series, we feed $x_i$ and the metadata of $x_j$ to the CIMP GAN and compare its output to the real reference image $x_j$, using SSIM~\citep{wang2004ssim}, PSNR, and LPIPS. As a baseline, we compare $x_i$ directly to $x_j$ with no model in the loop, indicating how similar two real images at different settings within the same series already are. Table~\ref{tab:style_transfer_pixel} shows that the SSIM, PSNR, and LPIPS values are very consistent between the CIMP GAN and the eval-series baseline across all three series, with the two rows differing by less than $0.02$ on SSIM, $1.3$ on PSNR, and $0.05$ on LPIPS in every case. This indicates that the CIMP GAN modifies the input by an amount comparable to the natural change between adjacent real acquisitions in the series. Among the three series, the CIMP GAN's similarity to the reference is highest for gain, suggesting the model captures the gain change most actively, while dwell shows the highest baseline similarity since consecutive images in the dwell series differ only modestly to begin with.

We also test distributional realism with the Kernel Inception Distance (KID)~\citep{binkowski2018demystifying}. Table~\ref{tab:style_transfer_kid} reports four comparisons. The first is CIMP GAN output against the real eval-set images, a direct realism check. The second is CIMP GAN output against the training corpus. The third is the 21 real eval-set images against the training corpus. This serves as a content-gap floor that isolates the distribution shift between the narrow eval-series content and the wider training corpus, independent of any model contribution. The fourth is two disjoint halves of a training pool against each other, a within-corpus sampling-noise floor. The last three comparisons are computed across five disjoint training-image pools of 500 images each with pinned seeds, so they share the same right-hand reference within each seed. Subset size is 21 for the first three comparisons, bounded by the eval-set size, and 50 for the noise floor.

The within-corpus floor ($3 \times 10^{-4}$) confirms that Inception features discriminate cleanly on this dataset. As expected, two random samples of real training images are indistinguishable. Most directly, the CIMP GAN's KID against the real eval-set images is near zero ($-0.016 \pm 0.009$), and we conclude that its output distribution almost perfectly matches the distribution of the real evaluation images. Against the training corpus, the CIMP GAN's KID ($0.185$) matches the content-gap floor ($0.189$) within one standard error. The common $\sim\!0.19$ value therefore reflects the shift between the narrow eval-series content and the wider training distribution, not a model artefact. Read together, Tables~\ref{tab:style_transfer_pixel} and~\ref{tab:style_transfer_kid} indicate that the CIMP GAN reproduces the experimental change at both the per-image and the distributional level, with the magnitude of induced change matching the natural intra-series variation and the output distribution matching the distribution of real eval-set images.

We qualitatively also compare the effect of changing dwell time, gain, and offset to effect of the same experimental changes on the microscope as shown in Figures~\ref{fig:exp_eval_dwell},~\ref{fig:exp_eval_gain}, and~\ref{fig:exp_eval_offset} in Appendix~\ref{app:evaluation_set}. As ground truth, we utilise the experimentally acquired evaluation set of gold particles on a carbon film recorded at five different dwell times, five different gains, and five different offsets. We choose to focus on these parameters as they are simple parameters to change experimentally, have a large impact on the image style, and are independent of other system components and thus do not affect the other metadata. In general, increasing dwell time does make the image smoother as expected, but also makes the background brighter, which does not happen in the experimental series. For gain, our model correctly predicts the behaviour of lowering and increasing brightness in the image. However, it does not successfully model the boundary condition where the experimental image is clipped. The CIMP-conditioned GAN model performs worst on the offset parameter; there is almost no difference between generated images across offset values, though the experimental series shows substantial change. This is likely due to limited diversity in the training set. 

We qualitatively evaluate style transfer for each metadata parameter individually, with the exception of pixel size. Since pixel size is determined by the image's field of view, altering it synthetically while keeping image dimensions fixed has no physical basis (further explained in Appendix~\ref{app:metadata_explanation}). The per-parameter qualitative trends, with the corresponding generator outputs, are reported in Appendix~\ref{app:evaluation_set}.

\subsection{Denoising}
\label{sec:denoising}

\begin{figure}[tbp]
    \centering
    \begin{subfigure}[t]{0.49\linewidth}
        \centering
        \begin{minipage}{0.32\linewidth}\centering
            \footnotesize\textbf{Raw}\\[2pt]
            \includegraphics[width=\linewidth]{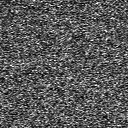}\\[1pt]
            \includegraphics[width=\linewidth]{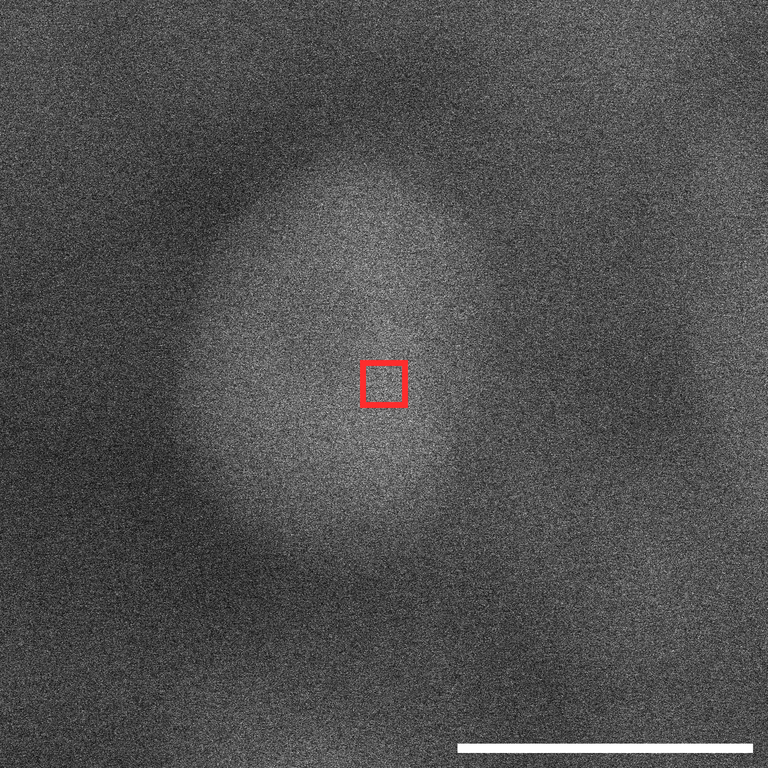}
        \end{minipage}\hfill
        \begin{minipage}{0.32\linewidth}\centering
            \footnotesize\textbf{N2V}\\[2pt]
            \includegraphics[width=\linewidth]{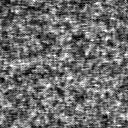}\\[1pt]
            \includegraphics[width=\linewidth]{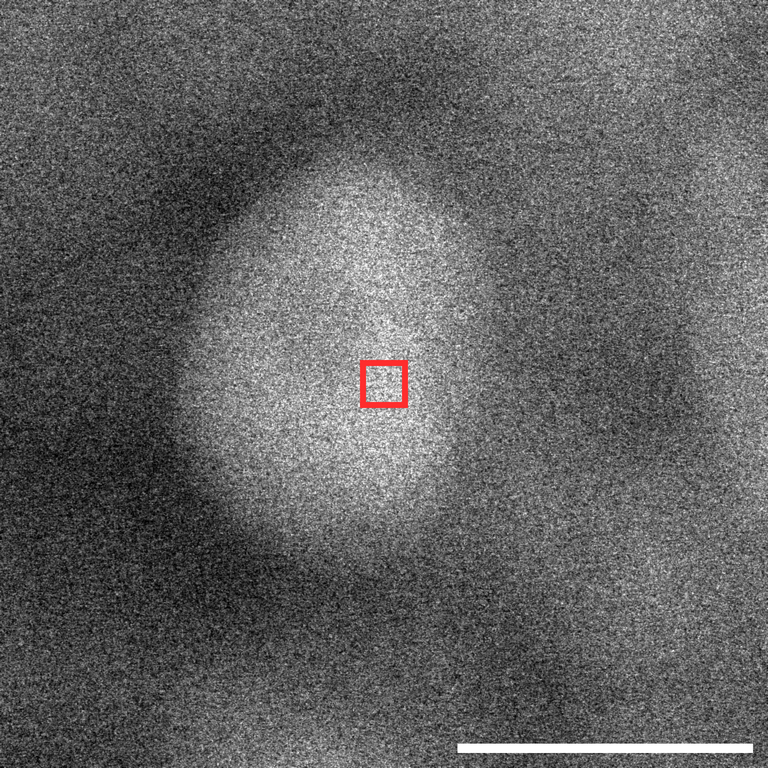}
        \end{minipage}\hfill
        \begin{minipage}{0.32\linewidth}\centering
            \footnotesize\textbf{CIMP GAN}\\[2pt]
            \includegraphics[width=\linewidth]{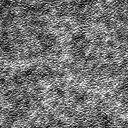}\\[1pt]
            \includegraphics[width=\linewidth]{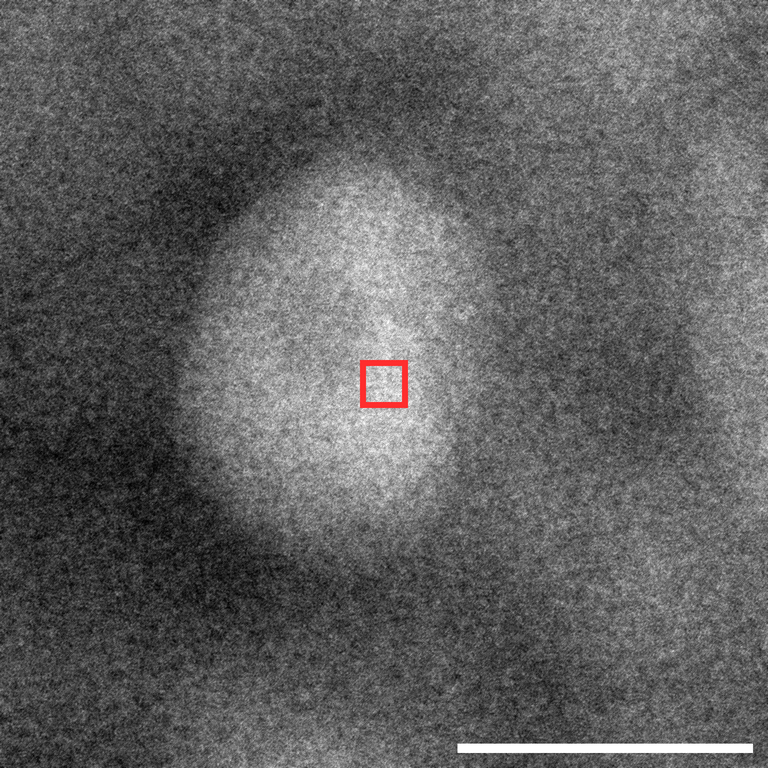}
        \end{minipage}
        \caption{Images of \ce{Pd/In/ZrO2} (in-situ \ce{CO2/H2}) with original settings of $t_{\text{dwell}}=100\,\text{ns}$, $I_{\text{beam}}=134.71\,\text{pA}$}
        \label{fig:zoom_grid_mat_a_10nm}
    \end{subfigure}\hfill
    \begin{subfigure}[t]{0.49\linewidth}
        \centering
        \begin{minipage}{0.32\linewidth}\centering
            \footnotesize\textbf{Raw}\\[2pt]
            \includegraphics[width=\linewidth]{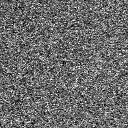}\\[1pt]
            \includegraphics[width=\linewidth]{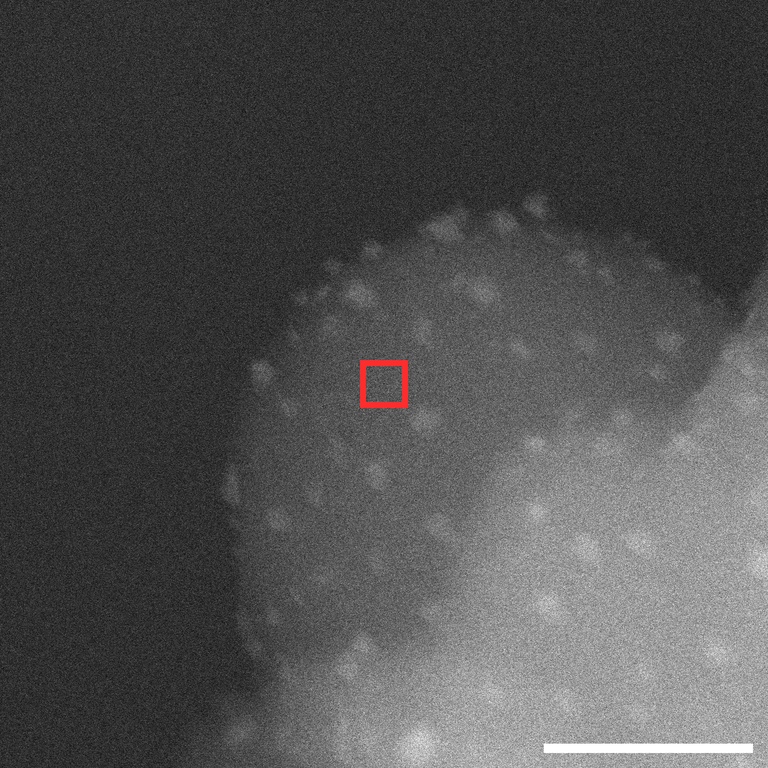}
        \end{minipage}\hfill
        \begin{minipage}{0.32\linewidth}\centering
            \footnotesize\textbf{N2V}\\[2pt]
            \includegraphics[width=\linewidth]{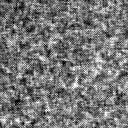}\\[1pt]
            \includegraphics[width=\linewidth]{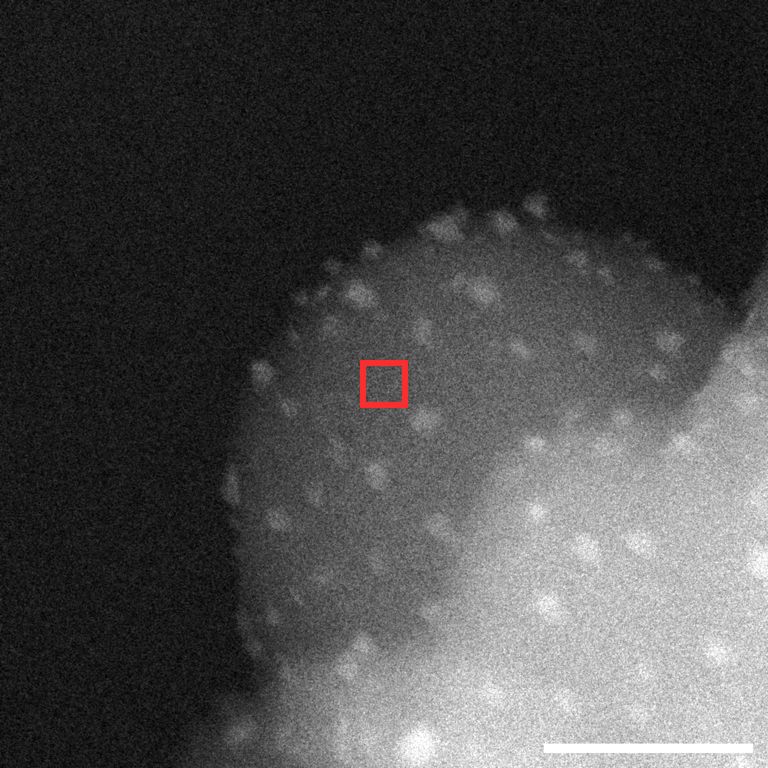}
        \end{minipage}\hfill
        \begin{minipage}{0.32\linewidth}\centering
            \footnotesize\textbf{CIMP GAN}\\[2pt]
            \includegraphics[width=\linewidth]{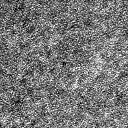}\\[1pt]
            \includegraphics[width=\linewidth]{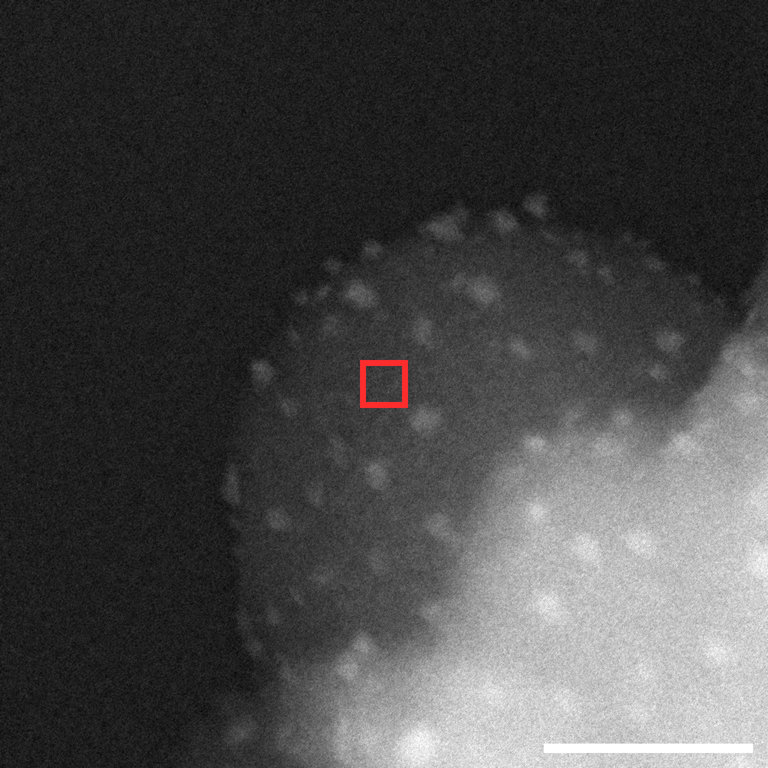}
        \end{minipage}
        \caption{Images of \ce{Au/TiO2} (in-situ) with original settings of $t_{\text{dwell}}=2\,\mu\text{s}$, $I_{\text{beam}}=24.34\,\text{pA}$}
        \label{fig:zoom_grid_mat_b_10nm}
    \end{subfigure}

    \vspace{6pt}

    \begin{subfigure}[t]{0.49\linewidth}
        \centering
        \begin{minipage}{0.32\linewidth}\centering
            \includegraphics[width=\linewidth]{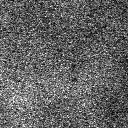}\\[1pt]
            \includegraphics[width=\linewidth]{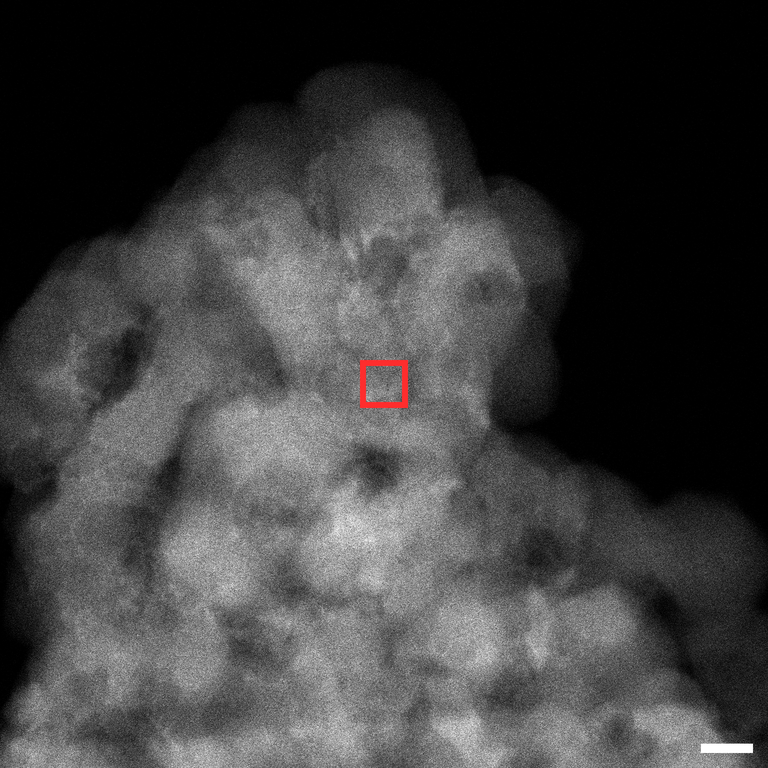}
        \end{minipage}\hfill
        \begin{minipage}{0.32\linewidth}\centering
            \includegraphics[width=\linewidth]{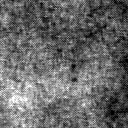}\\[1pt]
            \includegraphics[width=\linewidth]{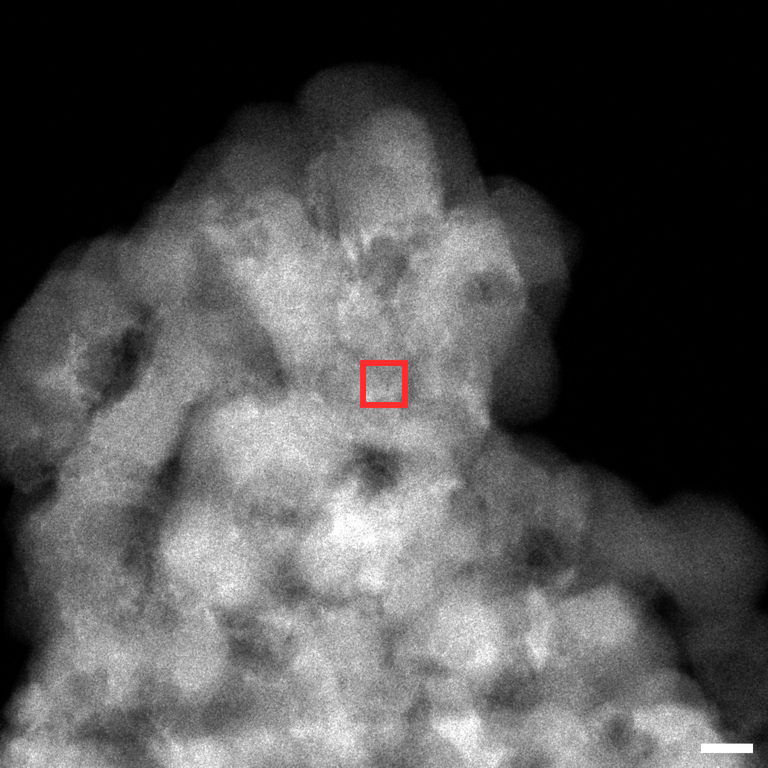}
        \end{minipage}\hfill
        \begin{minipage}{0.32\linewidth}\centering
            \includegraphics[width=\linewidth]{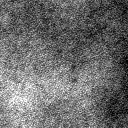}\\[1pt]
            \includegraphics[width=\linewidth]{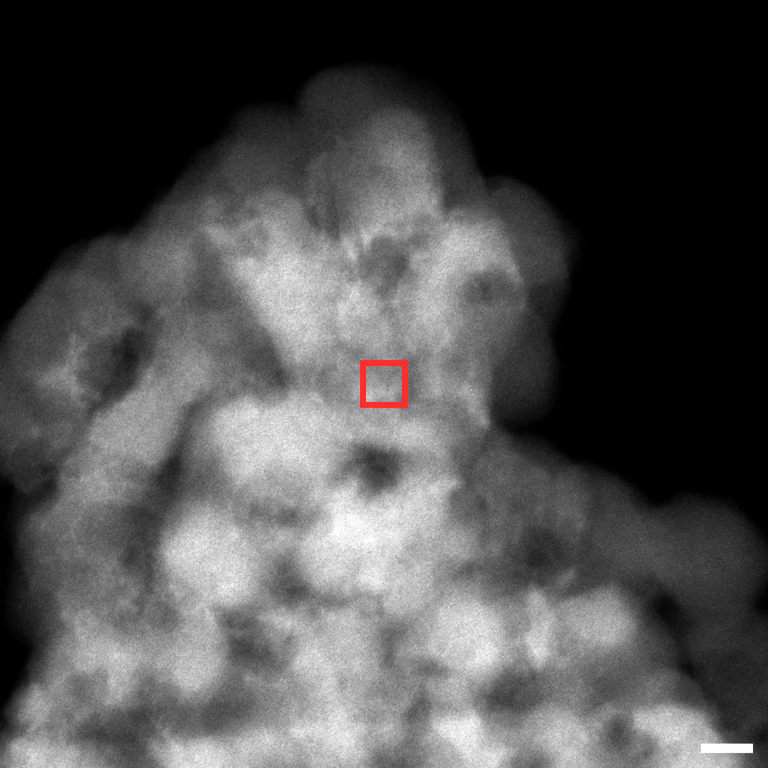}
        \end{minipage}
        \caption{Images of \ce{Ru/TiO2} with original settings of $t_{\text{dwell}}=2\,\mu\text{s}$, $I_{\text{beam}}=31.94\,\text{pA}$}
        \label{fig:zoom_grid_mat_c_10nm}
    \end{subfigure}\hfill
    \begin{subfigure}[t]{0.49\linewidth}
        \centering
        \begin{minipage}{0.32\linewidth}\centering
            \includegraphics[width=\linewidth]{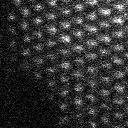}\\[1pt]
            \includegraphics[width=\linewidth]{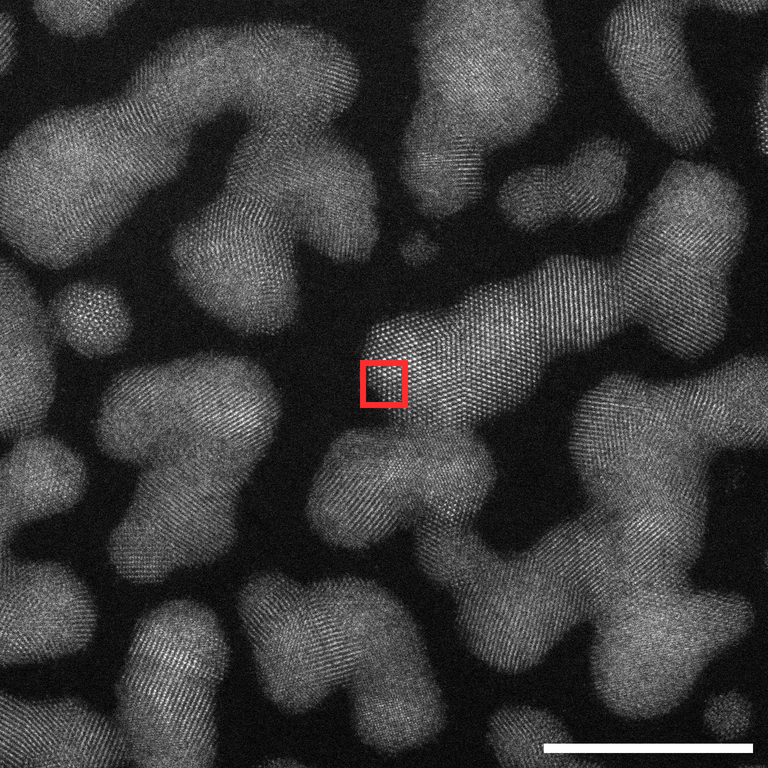}
        \end{minipage}\hfill
        \begin{minipage}{0.32\linewidth}\centering
            \includegraphics[width=\linewidth]{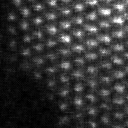}\\[1pt]
            \includegraphics[width=\linewidth]{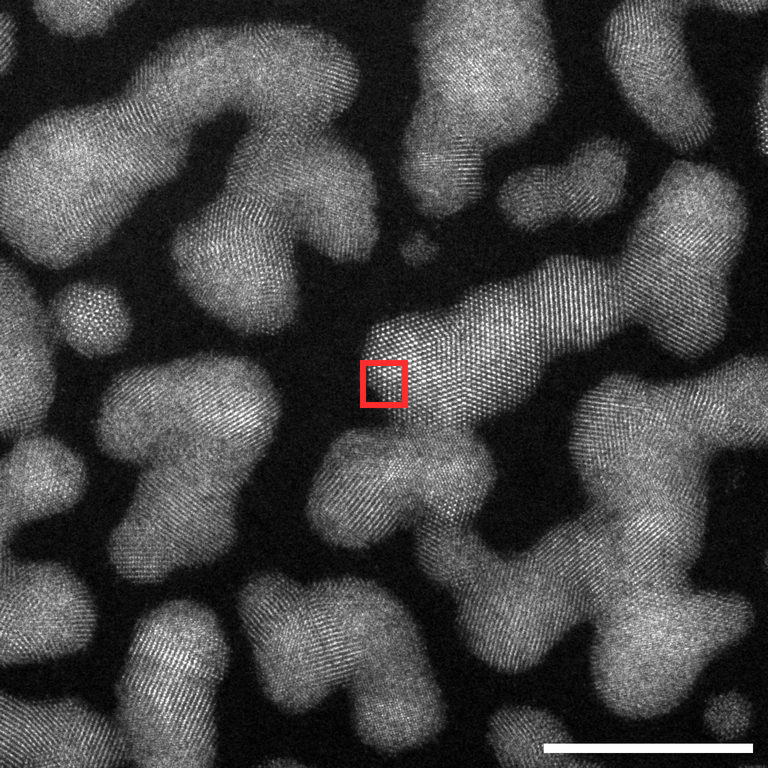}
        \end{minipage}\hfill
        \begin{minipage}{0.32\linewidth}\centering
            \includegraphics[width=\linewidth]{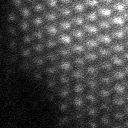}\\[1pt]
            \includegraphics[width=\linewidth]{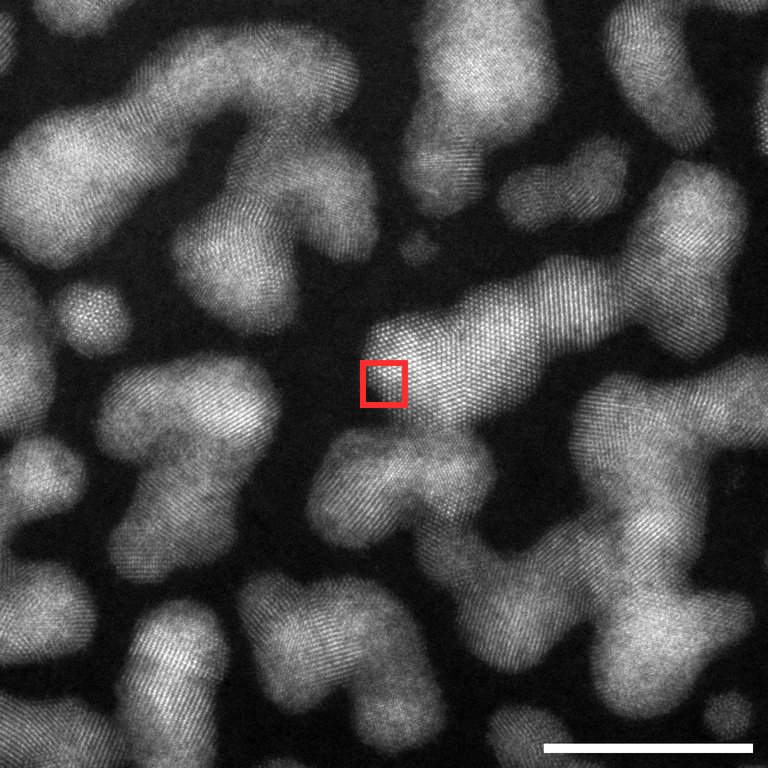}
        \end{minipage}
        \caption{Images of \ce{Ni}(100) with original settings of $t_{\text{dwell}}=2\,\mu\text{s}$, $I_{\text{beam}}=28.33\,\text{pA}$}
        \label{fig:zoom_grid_mat_d_10nm}
    \end{subfigure}

    \caption{Four denoising examples spanning distinct material systems. Within each panel, columns are raw / Noise2Void / CIMP-GAN; the top row is a $128\times128$ crop and the bottom row is the full $2048\times 2048$ frame. CIMP-GAN target metadata is $t_{\text{dwell}}=24.4\,\mu\text{s}$, $I_{\text{beam}}=86.95\,\text{pA}$ in all four panels. Scale bars on the full-frame images represent $10$ nm.}
    \label{fig:zoom_compare_grid_materials}
\end{figure}

Finally, we utilize the CIMP-conditioned GAN model as a denoiser by feeding it a target embedding $e_{\text{tgt}}$ derived from the source metadata with virtually increased dwell time and beam current. We compare our model to the Noise2Void implementation of ~\citet{Thornley2026} and to the GAN-based Noise2Atom denoiser~\citep{FengNoise2Atom}, both retrained on our dataset for a best performance comparison. 

Qualitatively, our CIMP GAN denoiser consistently outputs more realistic looking denoised images than Noise2Void, which is overly aggressive in its denoising, suffers from checkerboard artifacts (Figure~\ref{fig:zoom_compare_grid_materials}), and tends to hallucinate content in background regions by smoothing speckle noise into gaussians. 
Qualitative Noise2Atom outputs on four representative held-out frames are shown in Figure~\ref{fig:n2a_samples} (Appendix~\ref{app:n2a_samples}). Noise2Atom fails on the broader denoising task, with complete content collapse on any input whose content is not a clean atomic-resolution lattice. We therefore do not consider Noise2Atom a comparable denoiser and exclude it from the user study below.

Quantitatively, we compare the CIMP-conditioned GAN denoiser to Noise2Void and the GAN-based Noise2Atom denoiser on 100 held-out validation frames (Table~\ref{tab:denoise_metrics}). Our CIMP GAN preserves the most edge magnitude (edge energy ratio $0.79$). Noise2Void preserves edge location best (gradient correlation $0.69$) but smooths aggressively and retains only $36\%$ of total edge magnitude. Noise2Atom collapses image content onto a learned atomic prior, scoring poorly on both edge metrics.

The three frequency-band rows in Table~\ref{tab:denoise_metrics} show how each method acts as a function of spatial frequency. Noise2Void shows a simple low-pass profile, suppressing high-frequency power by roughly an order of magnitude while leaving low frequencies essentially unchanged. Our CIMP GAN suppresses power gently and almost uniformly across the spectrum, consistent with its denoising mechanism being a metadata-driven style shift rather than a frequency cutoff. Noise2Atom suppresses high-frequency content but adds power in the low-to-mid frequency range, reflecting the periodic atomic-lattice content the model imprints on every input regardless of the underlying sample. The full radial spectra are plotted in Figure~\ref{fig:psd} (Appendix~\ref{app:denoiser_psd}).

\begin{table}[H]
\centering
\caption{Reference-free denoising metrics computed across 100 held-out validation frames. The bottom three rows report $\log_{10}$ ratios of raw-to-denoised radial power averaged within each frequency band. Positive values indicate suppression, negative values indicate added power. The full metric set is in Table~\ref{tab:denoise_metrics_full} (Appendix~\ref{app:denoiser_psd}).}
\label{tab:denoise_metrics}
\footnotesize
\begin{tabular}{lccc}
\toprule
Metric & CIMP GAN (ours) & Noise2Void & Noise2Atom \\
\midrule
Gradient correlation~$\uparrow$                 & $0.57 \pm 0.02$ & $\mathbf{0.69 \pm 0.01}$    & $0.24 \pm 0.01$    \\
Edge energy ratio~$\uparrow$                    & $\mathbf{0.79 \pm 0.03}$ & $0.36 \pm 0.01$    & $0.58 \pm 0.07$    \\
\midrule
PSD ratio, low-freq ($\log_{10}$)                & $\mathbf{0.18 \pm 0.03}$ & $-0.012 \pm 0.001$ & $0.69 \pm 0.06$    \\
PSD ratio, mid-freq ($\log_{10}$)                & $\mathbf{0.24 \pm 0.03}$ & $0.44 \pm 0.02$    & $-0.09 \pm 0.03$   \\
PSD ratio, high-freq ($\log_{10}$)    & $\mathbf{0.25 \pm 0.03}$ & $1.08 \pm 0.02$    & $0.87 \pm 0.07$    \\
\bottomrule
\end{tabular}
\end{table}

\vspace{-20pt}

\begin{table}[H]
\centering
\caption{Per-rater counts of CIMP GAN choices in the user study, with the Wilson $95\%$ CI on the rate, and a two-sided binomial $p$-value against the $50/50$ null.}
\label{tab:user_study_prelim}
\footnotesize
\begin{tabular}{lrlr}
\toprule
Rater & Judgments & CIMP GAN chosen (Wilson $95\%$ CI) & Binomial $p$ vs $50/50$ \\
\midrule
Rater 1 & 200 & 137 (68.5\% [61.8, 74.5]) & $<10^{-3}$ \\
Rater 2 & 101 &  77 (76.2\% [67.1, 83.5]) & $<10^{-3}$ \\
Rater 3 &  76 &  65 (85.5\% [75.9, 91.7]) & $<10^{-3}$ \\
Rater 4 &  73 &  37 (50.7\% [39.5, 61.8]) & $1.00$ \\
Rater 5 &  36 &  25 (69.4\% [53.1, 82.0]) & $0.03$ \\
\midrule
Total   & 486 & 341 (70.2\% [66.0, 74.1]) & $<10^{-3}$ \\
\bottomrule
\end{tabular}
\end{table}

To confirm the quality of CIMP-GAN as a denoiser, we conduct a blind user study with 5 independent experimental microscopists. The user study presented 200 random (raw, Noise2Void, CIMP-GAN) triplets from the validation set, where each image was a $128\times128$ image patch and the CIMP-GAN images were conditioned with target metadata of $t_{\text{dwell}}=24.4\,\mu\text{s}$, $I_{\text{beam}}=86.95\,\text{pA}$. 
The full instructions as well as the judgment overlap between raters are reported in Appendix~\ref{app:user_study}.

Across all $486$ judgments, CIMP GAN is chosen on $70.2\%$ (Wilson $95\%$ CI $[66.0, 74.1]$) of trials, significantly above $50\%$ (binomial $p < 10^{-3}$). Per-rater preference is uneven, four raters prefer CIMP at $68$--$86\%$ while one is indistinguishable from chance at $50.7\%$ (Table~\ref{tab:user_study_prelim}). Coverage is uneven across raters, ranging from 36 to 200 judgments per rater. A general impression from some of the raters was that "the one with cross artefacts" (N2V) generates sharper but more unrealistic images, and works better when the signal of the raw image is already quite high.

\section{Limitations \& Conclusion}

We introduced CIMP, a CLIP-style contrastive framework that learns a joint embedding space over HAADF-STEM images and their acquisition metadata from 7{,}330 paired examples. The factored representation we obtain from contrastive image-metadata alignment is a foundation other workflows can condition on rather than infrastructure for any single application. It could anchor segmentation or analysis networks robust to varying instrument state, serve as a retrieval index over a metadata-tagged corpus, or act as a perceptual loss for a physics-informed denoiser, as shown.

Two principal limitations remain. Our dataset of 7{,}330 single-instrument images is modest by computer-vision standards, and the data underrepresents boundary conditions such as clipped detectors and saturated extremes that operators rarely record (full discussion in Appendix~\ref{app:limitations}). Future work will address both by extending the dataset across instruments, quantifying the denoiser against stronger baselines on paired data, and investigating how CIMP embeddings transfer to other analysis tasks.

More broadly, this work represents a step toward autonomous high-throughput microscopy, in which closed-loop instruments select, acquire, and interpret images with reduced human supervision. A factored representation grounded in physical microscope state is a foundational requirement for any such system, and the same representation extends to interactive workflows, which could provide operators with tools that reason over acquisition parameters in real time. However, the utility of such approaches scales with the diversity of paired corpora available to the field, and we hope this work motivates broader release of the metadata-rich data that remains underutilized in the electron microscopy community.

\bibliographystyle{plainnat}
\bibliography{references}

@article{chemistry7050160,
AUTHOR = {Ivanova, Nina M. and Kashin, Alexey S. and Ananikov, Valentine P.},
TITLE = {Lost Data in Electron Microscopy},
JOURNAL = {Chemistry},
VOLUME = {7},
YEAR = {2025},
NUMBER = {5},
ARTICLE-NUMBER = {160},
URL = {https://www.mdpi.com/2624-8549/7/5/160},
ISSN = {2624-8549},
DOI = {10.3390/chemistry7050160}
}

@article{PhysRevLett.102.096101,
  title = {Atomic-Resolution Imaging with a Sub-50-pm Electron Probe},
  author = {Erni, Rolf and Rossell, Marta D. and Kisielowski, Christian and Dahmen, Ulrich},
  journal = {Phys. Rev. Lett.},
  volume = {102},
  issue = {9},
  pages = {096101},
  numpages = {4},
  year = {2009},
  month = {Mar},
  publisher = {American Physical Society},
  doi = {10.1103/PhysRevLett.102.096101},
  url = {https://link.aps.org/doi/10.1103/PhysRevLett.102.096101}
}

@article{EGERTON2025103819,
title = {Two- and three-dimensional electron imaging of beam-sensitive specimens},
journal = {Micron},
volume = {194},
pages = {103819},
year = {2025},
issn = {0968-4328},
doi = {https://doi.org/10.1016/j.micron.2025.103819},
url = {https://www.sciencedirect.com/science/article/pii/S096843282500037X},
author = {R.F. Egerton}
}

@Inbook{Reimer1984,
author="Reimer, Ludwig",
title="Specimen Damage by Electron Irradiation",
bookTitle="Transmission Electron Microscopy: Physics of Image Formation and Microanalysis",
year="1984",
publisher="Springer Berlin Heidelberg",
address="Berlin, Heidelberg",
pages="421--453",
isbn="978-3-662-13553-2",
doi="10.1007/978-3-662-13553-2_10",
url="https://doi.org/10.1007/978-3-662-13553-2_10"
}

@article{FengNoise2Atom,
   author = {Wang, Feng and Henninen, Trond R. and Keller, Debora and Erni, Rolf},
   title = {Noise2Atom: unsupervised denoising for scanning transmission electron microscopy images},
   journal = {Applied Microscopy},
   volume = {50},
   number = {1},
   pages = {23},
   ISSN = {2287-4445},
   DOI = {10.1186/s42649-020-00041-8},
   url = {https://doi.org/10.1186/s42649-020-00041-8},
   year = {2020},
   type = {Journal Article}
}

@article{park_denoise,
author = {Joodeok Kim  and Jinho Rhee  and Sungsu Kang  and Mingyu Jung  and Jihoon Kim  and Miji Jeon  and Junsun Park  and Jimin Ham  and Byung Hyo Kim  and Won Chul Lee  and Soung-Hun Roh  and Jungwon Park },
title = {Self-supervised machine learning framework for high-throughput electron microscopy},
journal = {Science Advances},
volume = {11},
number = {14},
pages = {eads5552},
year = {2025},
doi = {10.1126/sciadv.ads5552},
URL = {https://www.science.org/doi/abs/10.1126/sciadv.ads5552},
eprint = {https://www.science.org/doi/pdf/10.1126/sciadv.ads5552}}

@Article{Thornley2026,
author={Thornley, William
and Sullivan-Allsop, Sam
and Cai, Rongsheng
and Clark, Nick
and Gorbachev, Roman
and Haigh, Sarah J.},
title={Noise2Void for denoising atomic resolution scanning transmission electron microscopy images},
journal={npj Computational Materials},
year={2026},
month={Jan},
day={13},
volume={12},
number={1},
pages={68},
issn={2057-3960},
doi={10.1038/s41524-025-01939-1},
url={https://doi.org/10.1038/s41524-025-01939-1}
}

@inproceedings{krull2019noise2void,
  author    = {Krull, Alexander and Buchholz, Tim-Oliver and Jug, Florian},
  title     = {Noise2Void --- Learning Denoising From Single Noisy Images},
  booktitle = {Proceedings of the IEEE/CVF Conference on Computer Vision and Pattern Recognition (CVPR)},
  pages     = {2129--2137},
  year      = {2019},
  doi       = {10.1109/CVPR.2019.00223}
}

@Article{Eliasson3D,
author={Eliasson, Henrik
and Wang, Fangjinhua
and Wang, Xi
and Barath, Daniel
and Pollefeys, Marc
and Erni, Rolf},
title={Morphology prediction of small nanoparticles in any orientation from single electron micrographs},
journal={npj Computational Materials},
year={2026},
month={May},
day={06},
issn={2057-3960},
doi={10.1038/s41524-026-02114-w},
url={https://doi.org/10.1038/s41524-026-02114-w}
}

@article{EliassonCycleGAN,
   author = {Eliasson, Henrik and Lothian, Angus and Surin, Ivan and Mitchell, Sharon and Pérez-Ramírez, Javier and Erni, Rolf},
   title = {Precise Size Determination of Supported Catalyst Nanoparticles via Generative AI and Scanning Transmission Electron Microscopy},
   journal = {Small Methods},
   pages = {2401108},
   ISSN = {2366-9608},
   DOI = {https://doi.org/10.1002/smtd.202401108},
   url = {https://onlinelibrary.wiley.com/doi/abs/10.1002/smtd.202401108},
   year = {2024},
   type = {Journal Article}
}

@InProceedings{pmlr-v80-lehtinen18a,
  title = 	 {{N}oise2{N}oise: Learning Image Restoration without Clean Data},
  author =       {Lehtinen, Jaakko and Munkberg, Jacob and Hasselgren, Jon and Laine, Samuli and Karras, Tero and Aittala, Miika and Aila, Timo},
  booktitle = 	 {Proceedings of the 35th International Conference on Machine Learning},
  pages = 	 {2965--2974},
  year = 	 {2018},
  editor = 	 {Dy, Jennifer and Krause, Andreas},
  volume = 	 {80},
  series = 	 {Proceedings of Machine Learning Research},
  month = 	 {10--15 Jul},
  publisher =    {PMLR},
  url = 	 {https://proceedings.mlr.press/v80/lehtinen18a.html}
}

@InProceedings{pmlr-v139-radford21a,
  title = 	 {Learning Transferable Visual Models From Natural Language Supervision},
  author =       {Radford, Alec and Kim, Jong Wook and Hallacy, Chris and Ramesh, Aditya and Goh, Gabriel and Agarwal, Sandhini and Sastry, Girish and Askell, Amanda and Mishkin, Pamela and Clark, Jack and Krueger, Gretchen and Sutskever, Ilya},
  booktitle = 	 {Proceedings of the 38th International Conference on Machine Learning},
  pages = 	 {8748--8763},
  year = 	 {2021},
  editor = 	 {Meila, Marina and Zhang, Tong},
  volume = 	 {139},
  series = 	 {Proceedings of Machine Learning Research},
  month = 	 {18--24 Jul},
  publisher =    {PMLR},
  url = 	 {https://proceedings.mlr.press/v139/radford21a.html}
}

@inproceedings{perez2018film,
  title={{FiLM}: Feature-wise Linear Modulation},
  author={Perez, Ethan and Strub, Florian and De Vries, Harm and Dumoulin, Vincent and Courville, Aaron},
  booktitle={Proceedings of the AAAI Conference on Artificial Intelligence},
  volume={32},
  number={1},
  year={2018}
}

@inproceedings{mao2017least,
  title={Least Squares Generative Adversarial Networks},
  author={Mao, Xudong and Li, Qing and Xie, Haoran and Lau, Raymond YK and Wang, Zhen and Smolley, Stephen Paul},
  booktitle={Proceedings of the IEEE International Conference on Computer Vision},
  pages={2794--2802},
  year={2017}
}

@inproceedings{zhang2018perceptual,
  title={The Unreasonable Effectiveness of Deep Features as a Perceptual Metric},
  author={Zhang, Richard and Isola, Phillip and Efros, Alexei A and Shechtman, Eli and Wang, Oliver},
  booktitle={Proceedings of the IEEE Conference on Computer Vision and Pattern Recognition},
  pages={586--595},
  year={2018}
}

@misc{lipman2022flow,
      title={Flow Matching for Generative Modeling}, 
      author={Yaron Lipman and Ricky T. Q. Chen and Heli Ben-Hamu and Maximilian Nickel and Matt Le},
      year={2022},
      eprint={2210.02747},
      archivePrefix={arXiv},
      primaryClass={cs.LG},
      url={https://arxiv.org/abs/2210.02747}
}

@misc{saharia2021palette,
      title={Palette: Image-to-Image Diffusion Models}, 
      author={Chitwan Saharia and William Chan and Huiwen Chang and Chris A. Lee and Jonathan Ho and Tim Salimans and David J. Fleet and Mohammad Norouzi},
      year={2021},
      eprint={2111.05826},
      archivePrefix={arXiv},
      primaryClass={cs.CV},
      url={https://arxiv.org/abs/2111.05826}
}

@misc{sytwu2023segmented,
      author       = {Sytwu, Katherine and Rangel DaCosta, Luis and Scott, Mary},
      title        = {Segmented high-resolution transmission electron microscopy images of nanoparticles},
      year         = {2023},
      publisher    = {Dryad},
      doi          = {10.7941/D1SP93},
      url          = {https://doi.org/10.7941/D1SP93},
      note         = {Dataset}
}

@misc{schwenker2020atomagined,
      author       = {Schwenker, Eric and Sen, Fatih and Wolverton, Chris and Ophus, Colin and Chan, Maria K. Y.},
      title        = {A Simulated Atomic-resolution HAADF STEM Imaging Dataset Containing Unique ICSD Structure Prototypes},
      year         = {2020},
      publisher    = {Materials Data Facility},
      doi          = {10.18126/szeq-yde5},
      url          = {https://doi.org/10.18126/szeq-yde5},
      note         = {Dataset}
}

@misc{eliasson2025synthetic,
      author       = {Eliasson, Henrik},
      title        = {Synthetic data from: Improving Nanoparticle Size Estimation from Scanning Transmission Electron Micrographs with a Multislice Surrogate Model},
      year         = {2025},
      publisher    = {Zenodo},
      doi          = {10.5281/zenodo.14608502},
      url          = {https://doi.org/10.5281/zenodo.14608502},
      note         = {Dataset}
}

@article{eliasson2025multislice,
      author       = {Eliasson, Henrik and Erni, Rolf},
      title        = {Improving Nanoparticle Size Estimation from Scanning Transmission Electron Micrographs with a Multislice Surrogate Model},
      journal      = {Nano Letters},
      volume       = {25},
      number       = {6},
      pages        = {2474--2479},
      year         = {2025},
      publisher    = {American Chemical Society},
      doi          = {10.1021/acs.nanolett.4c06025},
      url          = {https://doi.org/10.1021/acs.nanolett.4c06025}
}

@article{lin2021temimagenet,
      author       = {Lin, Ruoqian and Zhang, Rui and Wang, Chunyang and Yang, Xiao-Qing and Xin, Huolin L.},
      title        = {TEMImageNet training library and AtomSegNet deep-learning models for high-precision atom segmentation, localization, denoising, and deblurring of atomic-resolution images},
      journal      = {Scientific Reports},
      volume       = {11},
      number       = {1},
      pages        = {5386},
      year         = {2021},
      publisher    = {Springer Nature},
      doi          = {10.1038/s41598-021-84499-w},
      url          = {https://doi.org/10.1038/s41598-021-84499-w}
}

@article{sanchezfernandez2023cloome,
      author       = {Sanchez-Fernandez, Ana and Rumetshofer, Elisabeth and Hochreiter, Sepp and Klambauer, G\"unter},
      title        = {CLOOME: contrastive learning unlocks bioimaging databases for queries with chemical structures},
      journal      = {Nature Communications},
      volume       = {14},
      number       = {1},
      pages        = {7339},
      year         = {2023},
      publisher    = {Springer Nature},
      doi          = {10.1038/s41467-023-42328-w},
      url          = {https://doi.org/10.1038/s41467-023-42328-w}
}

@inproceedings{taleb2022contig,
      author       = {Taleb, Aiham and Kirchler, Matthias and Monti, Remo and Lippert, Christoph},
      title        = {ContIG: Self-supervised Multimodal Contrastive Learning for Medical Imaging with Genetics},
      booktitle    = {2022 IEEE/CVF Conference on Computer Vision and Pattern Recognition (CVPR)},
      pages        = {20876--20889},
      year         = {2022},
      publisher    = {IEEE},
      doi          = {10.1109/CVPR52688.2022.02024},
      url          = {https://doi.org/10.1109/CVPR52688.2022.02024}
}

@InProceedings{Sheth_2021_ICCV,
    author = {Sheth, Dev Yashpal and Mohan, Sreyas and Vincent, Joshua and Manzorro, Ramon and Crozier, Peter A. and Khapra, Mitesh M. and Simoncelli, Eero P. and Fernandez-Granda, Carlos},
    title = {Unsupervised Deep Video Denoising},
    booktitle = {Proceedings of the IEEE/CVF International Conference on Computer Vision (ICCV)},
    month = {October},
    year = {2021}
}

@article{Khan2023,
  author = {Khan, Abid and Lee, Chia-Hao and Huang, Pinshane Y. and Clark, Bryan K.},
  title = {Leveraging generative adversarial networks to create realistic scanning transmission electron microscopy images},
  journal = {npj Computational Materials},
  year = {2023},
  volume = {9},
  number = {1},
  pages = {85},
  month = {may},
  doi = {10.1038/s41524-023-01042-3},
  url = {https://doi.org/10.1038/s41524-023-01042-3},
  issn = {2057-3960}
}

@inproceedings{deng2009imagenet,
  author    = {Deng, Jia and Dong, Wei and Socher, Richard and Li, Li-Jia and Li, Kai and Fei-Fei, Li},
  title     = {{ImageNet}: A Large-Scale Hierarchical Image Database},
  booktitle = {IEEE Conference on Computer Vision and Pattern Recognition (CVPR)},
  year      = {2009},
  pages     = {248--255},
  doi       = {10.1109/CVPR.2009.5206848}
}

@article{Lobato2024,
  author = {Lobato, I. and Friedrich, T. and Van Aert, S.},
  title = {Deep convolutional neural networks to restore single-shot electron microscopy images},
  journal = {npj Computational Materials},
  year = {2024},
  volume = {10},
  number = {1},
  pages = {10},
  month = {jan},
  doi = {10.1038/s41524-023-01188-0},
  url = {https://doi.org/10.1038/s41524-023-01188-0},
  issn = {2057-3960}
}

@inproceedings{binkowski2018demystifying,
  author    = {Bi{\'n}kowski, Miko{\l}aj and Sutherland, Danica J. and Arbel, Michael and Gretton, Arthur},
  title     = {Demystifying {MMD GAN}s},
  booktitle = {International Conference on Learning Representations (ICLR)},
  year      = {2018}
}

@article{sawada2015si,
  author  = {Sawada, Hidetaka and Shimura, Naoki and Hosokawa, Fumio and Shibata, Naoya and Ikuhara, Yuichi},
  title   = {Resolving 45-pm-separated {Si--Si} atomic columns with an aberration-corrected {STEM}},
  journal = {Microscopy},
  volume  = {64},
  number  = {3},
  pages   = {213--217},
  year    = {2015},
  doi     = {10.1093/jmicro/dfv014},
  url     = {https://doi.org/10.1093/jmicro/dfv014}
}

@article{krivanek2015aberration,
  author  = {Krivanek, O. L. and Chisholm, M. F. and Dellby, N. and Murfitt, M. F.},
  title   = {Aberration-corrected {STEM} for atomic-resolution imaging and analysis},
  journal = {Journal of Microscopy},
  volume  = {259},
  number  = {3},
  pages   = {165--172},
  year    = {2015},
  doi     = {10.1111/jmi.12254},
  url     = {https://doi.org/10.1111/jmi.12254}
}

@inproceedings{ronneberger2015unet,
  title={U-Net: Convolutional Networks for Biomedical Image Segmentation},
  author={Ronneberger, Olaf and Fischer, Philipp and Brox, Thomas},
  booktitle={Medical Image Computing and Computer-Assisted Intervention -- MICCAI 2015},
  pages={234--241},
  year={2015},
  publisher={Springer International Publishing},
  doi={10.1007/978-3-319-24574-4_28}
}

@inproceedings{zhu2017cyclegan,
  author    = {Zhu, Jun-Yan and Park, Taesung and Isola, Phillip and Efros, Alexei A.},
  title     = {Unpaired Image-to-Image Translation using Cycle-Consistent Adversarial Networks},
  booktitle = {Proceedings of the IEEE International Conference on Computer Vision (ICCV)},
  pages     = {2223--2232},
  year      = {2017},
  doi       = {10.1109/ICCV.2017.244}
}

@article{oord2018cpc,
  author  = {van den Oord, Aaron and Li, Yazhe and Vinyals, Oriol},
  title   = {Representation Learning with Contrastive Predictive Coding},
  journal = {arXiv preprint arXiv:1807.03748},
  year    = {2018},
  url     = {https://arxiv.org/abs/1807.03748}
}

@inproceedings{he2016resnet,
  author    = {He, Kaiming and Zhang, Xiangyu and Ren, Shaoqing and Sun, Jian},
  title     = {Deep Residual Learning for Image Recognition},
  booktitle = {Proceedings of the IEEE Conference on Computer Vision and Pattern Recognition (CVPR)},
  pages     = {770--778},
  year      = {2016},
  doi       = {10.1109/CVPR.2016.90}
}

@inproceedings{dosovitskiy2021vit,
  author    = {Dosovitskiy, Alexey and Beyer, Lucas and Kolesnikov, Alexander and Weissenborn, Dirk and Zhai, Xiaohua and Unterthiner, Thomas and Dehghani, Mostafa and Minderer, Matthias and Heigold, Georg and Gelly, Sylvain and Uszkoreit, Jakob and Houlsby, Neil},
  title     = {An Image is Worth 16x16 Words: Transformers for Image Recognition at Scale},
  booktitle = {International Conference on Learning Representations (ICLR)},
  year      = {2021},
  url       = {https://openreview.net/forum?id=YicbFdNTTy}
}

@article{vandermaaten2008tsne,
  author  = {van der Maaten, Laurens and Hinton, Geoffrey},
  title   = {Visualizing Data using t-SNE},
  journal = {Journal of Machine Learning Research},
  volume  = {9},
  pages   = {2579--2605},
  year    = {2008},
  url     = {https://www.jmlr.org/papers/v9/vandermaaten08a.html}
}

@article{wang2004ssim,
  author  = {Wang, Zhou and Bovik, Alan C. and Sheikh, Hamid R. and Simoncelli, Eero P.},
  title   = {Image Quality Assessment: From Error Visibility to Structural Similarity},
  journal = {IEEE Transactions on Image Processing},
  volume  = {13},
  number  = {4},
  pages   = {600--612},
  year    = {2004},
  doi     = {10.1109/TIP.2003.819861}
}

@misc{abolhasani2026autonomous,
  title        = {On the Need for Autonomous Science Instruments: A Call to Action},
  author       = {Abolhasani, Milad and Ahmadi, Mahshid and Baird, Sterling and Berlinguette, Curtis P. and Brown, Keith A. and Fenning, David and Foster, Ian T. and Gottstein, Willi and Pablo-Garcia, Sergio and Gomes, Gabe and Hattrick-Simpers, Jason R. and Hein, Jason E. and Hitosugi, Taro and Kalinin, Sergei V. and Maruyama, Benji and Schrier, Joshua and Seifrid, Martin and Sparks, Taylor D. and Sun, Shijing and Sutherland, Brandon R. and Szymanski, Nathan J. and Thomas, Dean and Walters, Lauren and Xu, Jie and Zeng, Yan and Yang, Charles},
  year         = {2026},
  month        = feb,
  howpublished = {ChemRxiv preprint},
  doi          = {10.26434/chemrxiv.10001836/v1},
  url          = {https://chemrxiv.org/doi/full/10.26434/chemrxiv.10001836/v1},
  note         = {Posted 02 February 2026}
}


\appendix

\section{Brief explanation of metadata parameters}
\label{app:metadata_explanation}

\begin{enumerate}
    \item \textbf{Pixel Size:} This is the physical size of a pixel (typically around 15-50 pm for high resolution imaging), and the lateral step between probe positions in the STEM raster scan. Changing the pixel size is the result of changing magnification, which in normal operation would result in a larger field of view. It does not make physical sense to ask a model to change pixel size of the image while keeping the image size fixed. Therefore it is ignored and also not shuffled in the metadata shuffling within batches nor in the validation set for evaluation.
    \item \textbf{Dwell Time:} Dwell time determines how long the electron beam stays at each pixel position. A longer dwell time leads to more electrons hitting the sample at each pixel position and consequently leads to a higher signal image. A shorter dwell time leads to reduced signal and a significant increase of poisson noise.
    \item \textbf{Beam Convergence Angle:} The convergence angle of the beam is the incident angle of the convergent STEM probe as it hits the sample. The convergence angle is a key parameter of the diffraction-limited achievable resolution of the microscope. Changing the convergence angle can be done in multiple ways, but in this dataset, it was mostly controlled by the size of an aperture.
    \item \textbf{Beam Current:} This is the current of the electron probe, as measured by the fluorescent viewing screen, which is placed after the sample and HAADF detector. While a larger beam current indicates that more electrons hit the sample, the effect on signal strength depends on how many electrons hit the HAADF detector which is affected by other metadata like inner collection angle described below. Beam current should in general be kept as low as possible to avoid damaging the sample and inducing phenomena that do not occur naturally.
    \item \textbf{Detector Gain:} The ``Gain'' or ``Contrast'' is a parameter that determines how much the signal on the detector is amplified.
    \item \textbf{Detector Offset:} The ``Offset'' or ``Brightness'' is a parameter that sets the black level of the detector reading.
    \item \textbf{Inner Collection Angle:} The HAADF detector is annular, and the inner collection angle defines the minimum scattering angle of electrons that contribute to the recorded signal. This parameter is important both because it determines the fraction of the incident beam collected and because it governs the imaging contrast mechanism. At high scattering angles, the signal is dominated by incoherent Rutherford scattering, which forms the basis of HAADF-STEM contrast and gives rise to the approximate $Z^{1.7}$ dependence and mass--thickness contrast. At lower collection angles, coherently scattered electrons are also detected, introducing diffraction contrast and altering the imaging mechanism. For HAADF-STEM, including the majority of images in our dataset, the inner collection angle is typically chosen to be at least $\sim 3$ times the probe convergence semi-angle.
\end{enumerate}

\section{Dataset Construction \& Contents}
\label{app:dataset_construction}
The CIMP dataset can be found on Zenodo:  \href{https://zenodo.org/records/20058874?preview=1&token=eyJhbGciOiJIUzUxMiJ9.eyJpZCI6I[...]V8yOtDxd10MiHQlLABDz0oj_1qaayfQaJUCRNYHBU052oDmAaKe2xHmOdBtTprA}{HERE}.

\begin{figure}[H]
    \centering
    \includegraphics[width=\linewidth]{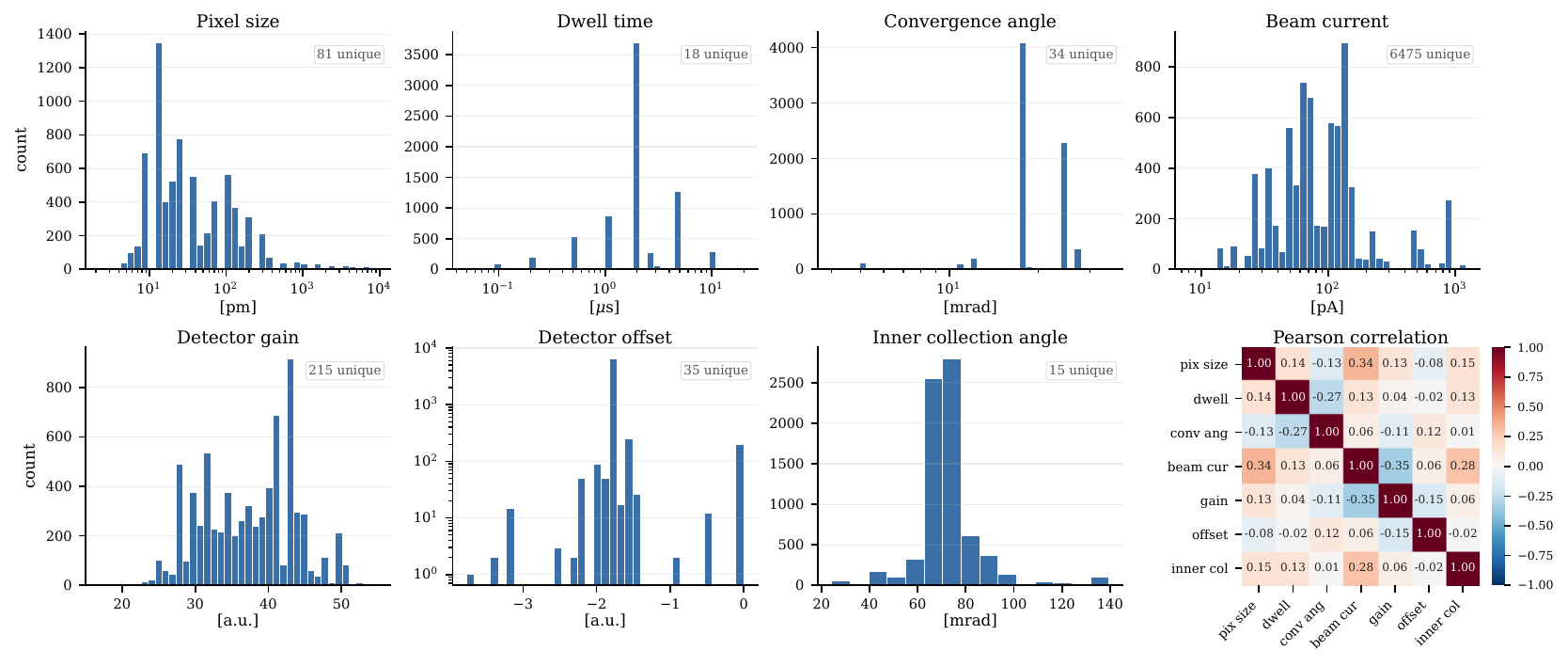}
    \caption{Distribution of acquisition metadata across 7,330 HAADF-STEM images.}
    \label{fig:metadata_distribution}
\end{figure}

Figure~\ref{fig:metadata_distribution} highlights several structural properties of the dataset that are relevant for downstream modelling. Pixel size, dwell time, and inner collection angle each take a small number of discrete values (81, 18, and 15 unique values respectively) as operators sweep these parameters through a fixed grid of magnifications, exposure presets, and detector geometries rather than varying them continuously. Beam current and detector gain are by contrast effectively continuous (thousands and hundreds of unique values). The many values for beam current originate from factors such as drift and degradation of the electron gun, lenses, and aperture placements. Convergence angle is bimodal, with two strongly populated values corresponding to the two aperture sizes routinely used during acquisition. The Pearson correlation panel (bottom-right) shows that no pair of parameters is strongly collinear, so each metadata dimension carries an independent supervisory signal for the contrastive loss. 

\paragraph{Distribution across material systems.} Material identity is recorded in the parent folder name of each source \texttt{.emd} file (the operator's directory hierarchy at acquisition time) rather than as an explicit metadata column. Pattern-matching the 130 folder names against catalyst-naming conventions assigns 5{,}017 of the 7{,}330 acquisitions ($\sim\!68\%$) to a clearly identifiable material family, while the remaining $\sim\!32\%$ lie in folders with non-material names (alignment shots, generic ``good''/``selected'' bins, date-only folders, or QC frames) and contribute valid HAADF--STEM image-metadata pairs but cannot be resolved to a specific material from filenames alone. Table~\ref{tab:material_breakdown} reports the resulting breakdown. The corpus is dominated by heterogeneous catalysts of practical interest: the single largest class is \ce{Pt/CeO2} ($\sim\!21\%$ of the dataset) with substantial additional coverage of \ce{CO2}-conversion catalysts, CO-oxidation systems, single-atom catalysts, and bare oxide supports. Many catalyst classes are sampled in multiple states (fresh, post-pretreatment, used, and in-situ under reactive gas), exposing the model to the structural variation a practitioner would expect across a typical experimental campaign. The family assignment in Table~\ref{tab:material_breakdown} is heuristic, based solely on operator-recorded folder names, and should be read as an indicative breakdown rather than a curated label set. 

\begin{table}[H]
\centering
\caption{Heuristic material-family breakdown of the 7{,}330 acquisitions, derived by pattern-matching the parent folder name of each source \texttt{.emd} file against catalyst-naming conventions. The ``Unspecified'' row covers folders with alignment, QC, or date-only names that do not encode the material in the path.}
\label{tab:material_breakdown}
\footnotesize
\setlength{\tabcolsep}{4pt}
\renewcommand{\arraystretch}{1.05}
\begin{tabular}{p{4.6cm}rr}
\toprule
\textbf{Material family} & \textbf{Imgs} & \textbf{\%} \\
\midrule
\ce{Pt/CeO2}                              & 1511 & 20.6 \\
Cr-oxide on \ce{CeO2}/\ce{Nb2O5}          & 153  & 2.1  \\
\ce{Pt/ZrO2}, \ce{Pt/TiO2}                & 37   & 0.5  \\
Pd/(In, Zn, Zr)-Ox                        & 538  & 7.3  \\
\ce{In2O3}-based (In/Zr-, In/Ti-, In/Hf-Ox) & 445 & 6.1 \\
Cu/Zn/Zr-, Cu/Zr-Ox (CZA family)          & 257  & 3.5  \\
Zn/Zr-Ox                                  & 56   & 0.8  \\
\ce{Ru/TiO2}                              & 228  & 3.1  \\
Ru/Ni \& Ni/Ru bimetallic                  & 98   & 1.3  \\
Ru-based (other / unspec.\ support)        & 173  & 2.4  \\
\ce{Au/TiO2}                               & 330  & 4.5  \\
Au/MnO\(_x\)                               & 266  & 3.6  \\
\ce{Au/CeO2}                               & 46   & 0.6  \\
Mn-/Ni- on \ce{CeO2} or other oxides       & 110  & 1.5  \\
Single- / few-atom on \ce{Al2O3} or \ce{CeO2} & 96 & 1.3  \\
Bare \ce{CeO2} model supports              & 273  & 3.7  \\
Bare \ce{TiO2}, \ce{HfO2}, aerogel supports & 148 & 2.0  \\
High-entropy oxide                         & 1    & 0.0  \\
Synthesis / pretreatment series            & 251  & 3.4  \\
\midrule
Unspecified (alignment, QC, date-only)     & 2311 & 31.5 \\
Unmatched                                  & 2    & 0.0  \\
\midrule
\textbf{Total}                             & \textbf{7330} & \textbf{100.0} \\
\bottomrule
\end{tabular}
\end{table}

\paragraph{Train/validation split strategy.} The 6{,}597 / 733 split ($\approx 90/10$) is stratified at the source-acquisition level rather than at the patch level. Specifically, source acquisitions (one \texttt{.emd} file per record) are randomly shuffled with a fixed seed and partitioned into train and validation sets; all $256\times256$ training crops sampled from a given source acquisition are then assigned to the same split. Because each of the 7{,}330 records corresponds to a distinct full-resolution image, this is equivalent to a $90/10$ split over images and prevents patches from the same image from leaking from training into validation, which would otherwise inflate retrieval metrics through near-duplicate matching of within-image texture.

\paragraph{Metadata extraction, provenance \& preprocessing} Images are stored in Thermo Fisher Scientific's Velox \texttt{.emd} (HDF5-based) format, and the seven acquisition-metadata fields are read directly from the per-image EMD record at dataset-construction time.

The four metadata dimensions that span several orders of magnitude (pixel size, dwell time, convergence angle, beam current) are stored in $\log_{10}$ space; all seven dimensions are additionally z-scored before being passed to the metadata encoder.

\section{Qualitative Analysis} 
\label{app:evaluation_set}
\begin{figure}[H]
    \centering
    \includegraphics[width=\linewidth]{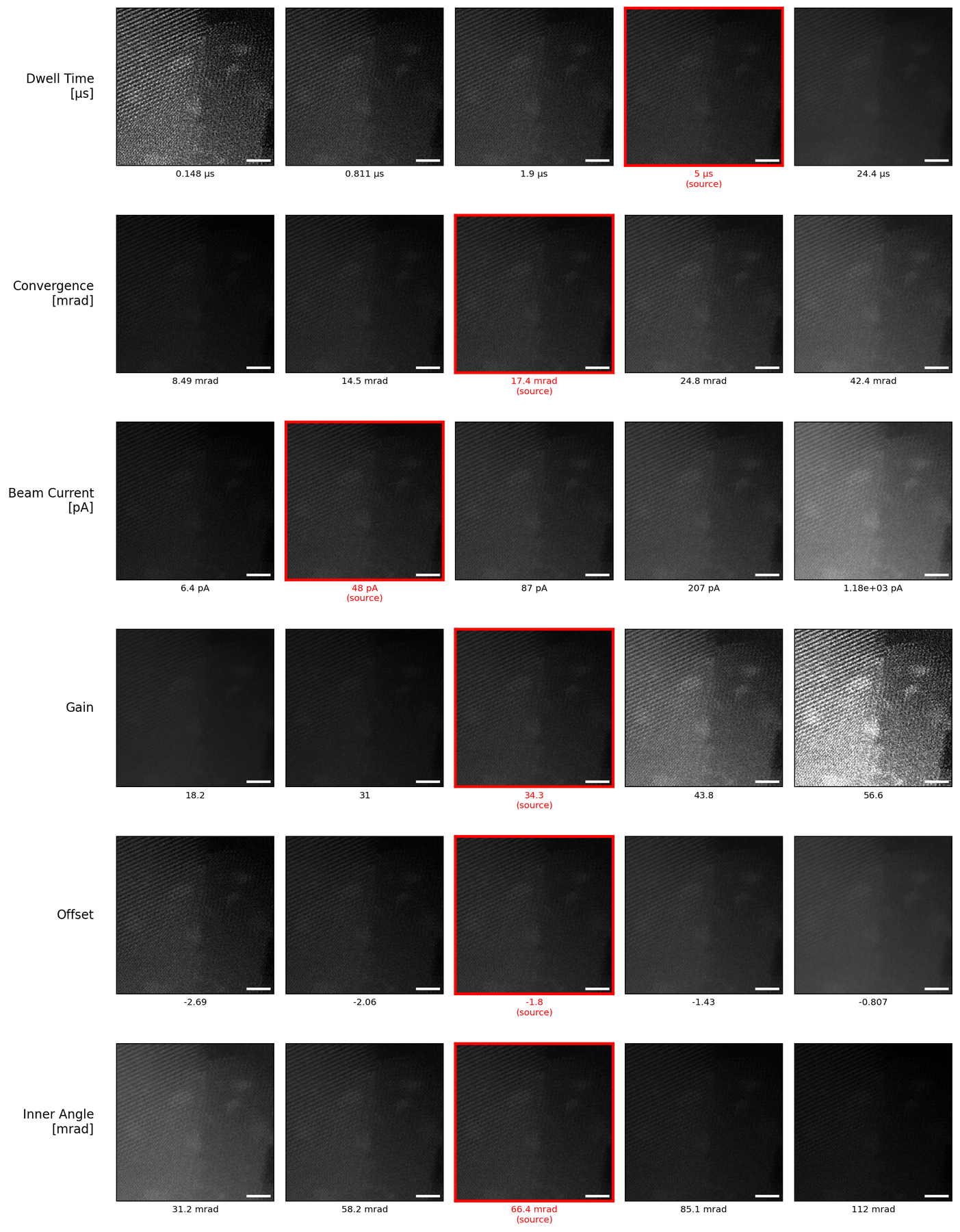}
    \caption{Our CIMP GAN style transfer model synthetically varying the metadata parameters of a validation image, one by one. Style-transferred images were generated at five target values ($-3\sigma$, $-1\sigma$, $0$, $+1\sigma$, and $+3\sigma$ of the distribution of that metadata in the full dataset). The image in the red frame is the original input image and it replaces whichever of the generated images that it has the closest metadata to. Collecting the metadata of all images with a red frames gives you the original metadata. \textit{Note: Gain and Offset are also denoted as Contrast and Brightness by some other STEM instruments/software, and their ranges can be different from Gain and Offset.} Scale bars are 2 nm.}
    \label{fig:sweeps}
\end{figure}

\begin{figure}[p]
    \centering
    \begin{subfigure}{\linewidth}
        \centering
        \includegraphics[width=0.85\linewidth]{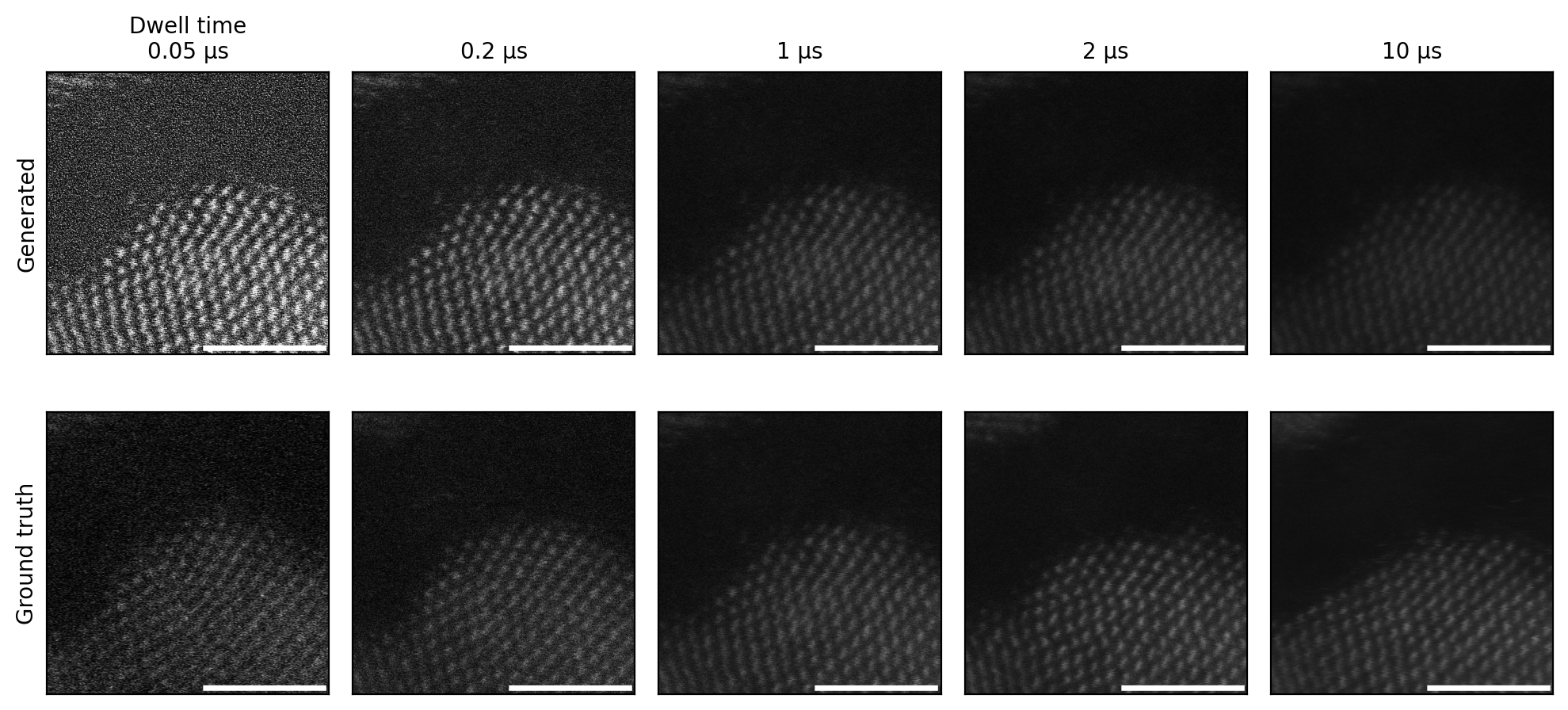}
        \caption{Results of varying the dwell time synthetically (top row) with our style transfer model and experimentally (bottom row). The expected behaviour of increasing the dwell time is a smoother image with less poisson noise but with a maintained pixel intensity distribution. We observe that our model indeed makes the image smoother, but also alters the intensity distribution by making background slightly brighter and object slightly darker. Furthermore, reducing dwell time virtually does not add as much noise as we observe experimentally.}
        \label{fig:exp_eval_dwell}
    \end{subfigure}

    \vspace{1ex}

    \begin{subfigure}{\linewidth}
        \centering
        \includegraphics[width=0.85\linewidth]{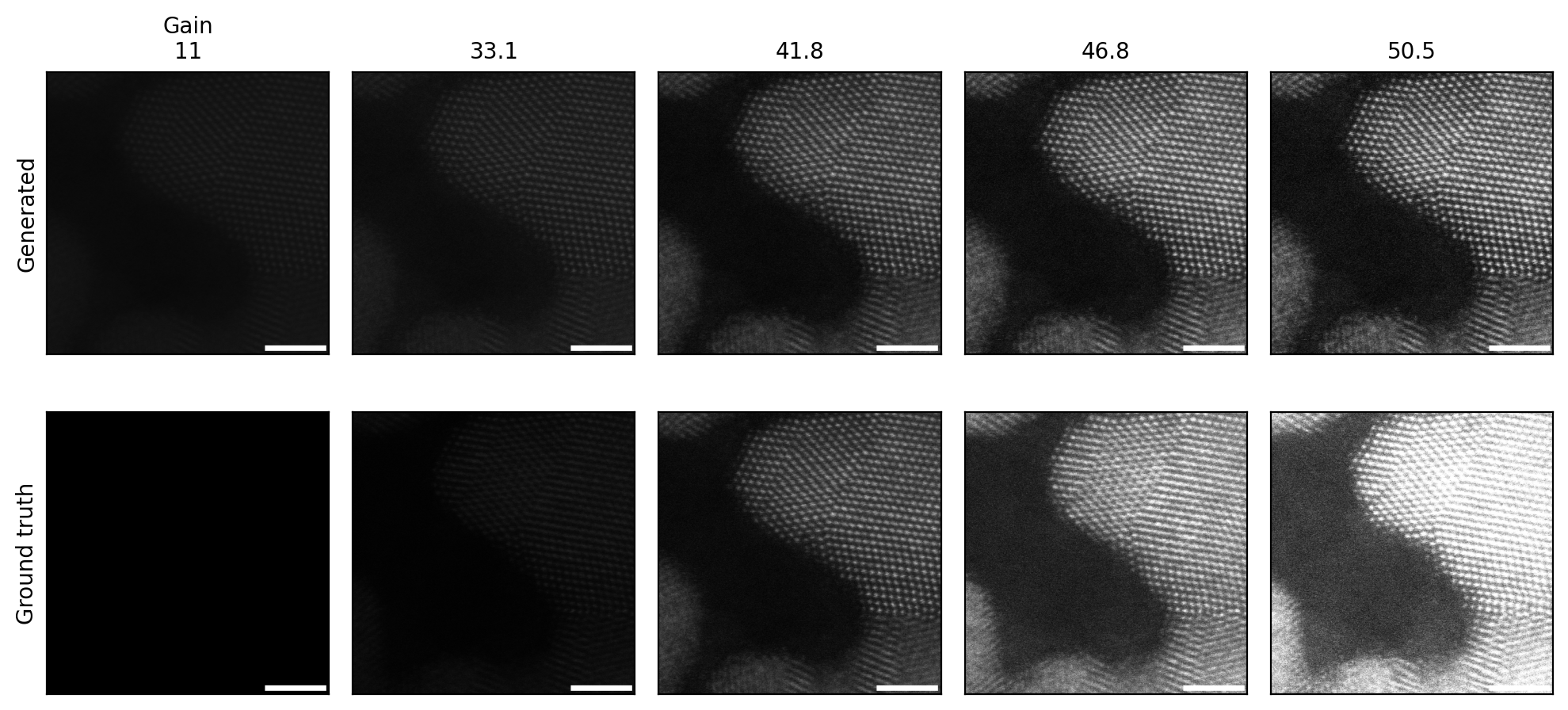}
        \caption{Results of varying the gain (or contrast) synthetically (top row) with our style transfer model and experimentally (bottom row). The expected behaviour of increasing the gain is an amplification of both signal and noise and a brighter image. We observe that our model can successfully model this trend but fails to model the upper bound where the the image would be clipped (further discussed in Appendix~\ref{app:limitations}).}
        \label{fig:exp_eval_gain}
    \end{subfigure}

    \vspace{1ex}

    \begin{subfigure}{\linewidth}
        \centering
        \includegraphics[width=0.85\linewidth]{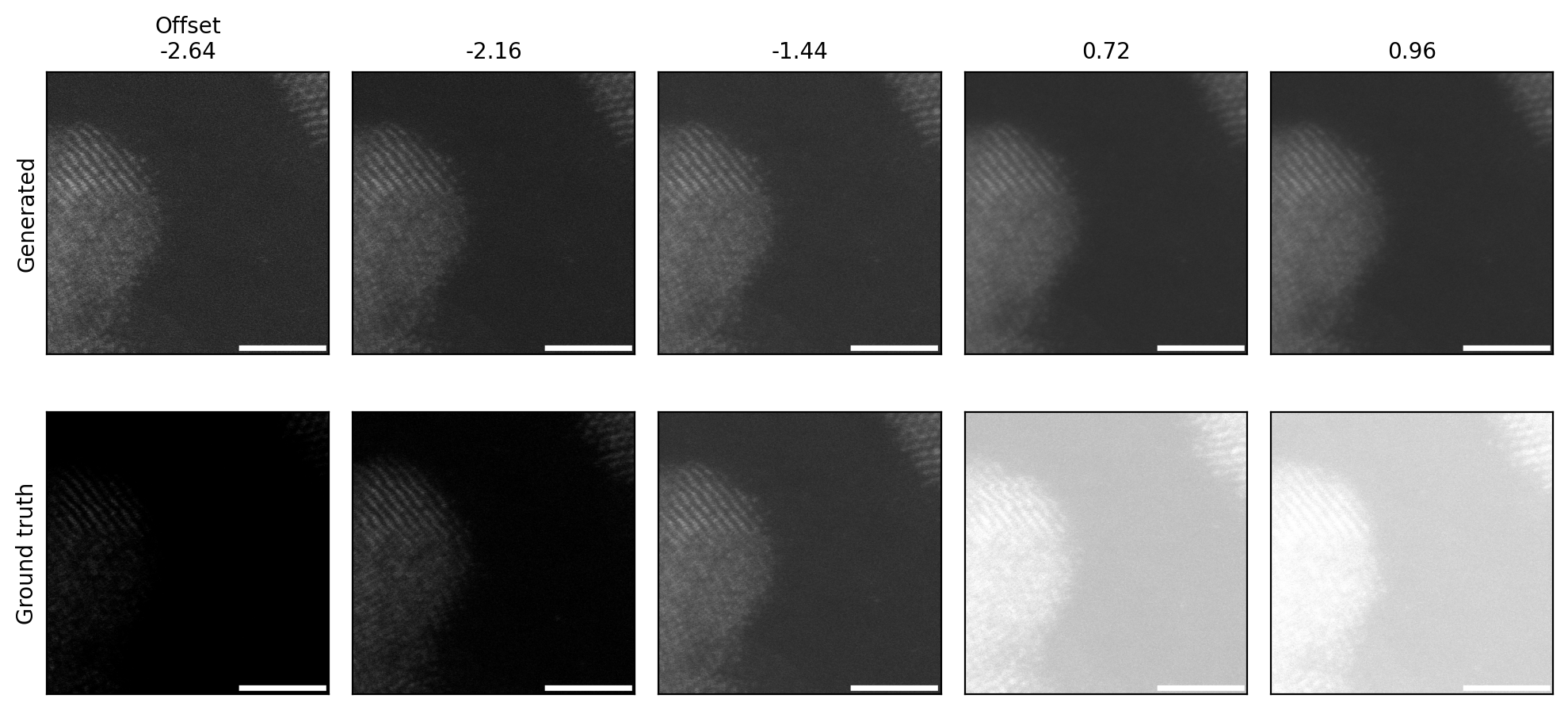}
        \caption{Results of varying the offset (or brightness) synthetically (top row) with our style transfer model and experimentally (bottom row). The expected behaviour of increasing the offset on this instrument is that the black level is lowered, pushing the visible content into a smaller part of the dynamic range, eventually clipping the image. Decreasing the offset leads to real signal being treated as black and parts of the object disappearing. We observe that a virtual change of offset does not closely mimic the behaviour of experimentally changing offset, we discuss possible reasons for this in Appendix~\ref{app:limitations}.}
        \label{fig:exp_eval_offset}
    \end{subfigure}
    \caption{Comparison between changing dwell time, gain, and offset virtually and experimentally. Scale bars are 2 nm.}
\end{figure}

\paragraph{Per-parameter qualitative trends.} For each parameter, Figure~\ref{fig:sweeps} shows the generator's output at five target values ($-3\sigma$, $-1\sigma$, $0$, $+1\sigma$, and $+3\sigma$ of the distribution of that metadata in the entire dataset), with all other metadata held at the source image's original values. For dwell time, gain, and offset, the model outputs that inform this analysis can additionally be seen in Figures~\ref{fig:exp_eval_dwell},~\ref{fig:exp_eval_gain}, and~\ref{fig:exp_eval_offset}. The general observed trends for each metadata are as follows:
\begin{enumerate}
    \item \textbf{Dwell time}: From low to high dwell time, the output image goes from noisy to smooth as expected.
    \item \textbf{Convergence angle}: When virtually increasing the convergence angle, the resulting effect is a brighter image. This is not what you would expect in theory, and likely happens because of the way convergence angle was controlled in the recording of the dataset. For the vast majority of images (see Figure~\ref{fig:metadata_distribution}), the convergence angle was set to two different values, controlled by changing the size of an aperture. A smaller aperture gave a smaller convergence angle but also blocked more of the electron beam, leading to a lower beam current which leads to a darker image. This bias is likely what is picked up by the model.
    \item \textbf{Beam Current}: For beam current, the trend is clear, more electrons lead to a brighter image. Eventually, this should also lead to clipping of the image, but as discussed in Appendix~\ref{app:limitations}, that boundary condition is not represented in the training data.
    \item \textbf{Gain}: Increased gain also leads to a brighter image. However, in contrast to the beam current sweep, the noise is also amplified.
    \item \textbf{Offset}: There is no clear trend for offset except that the background appears slightly darker for lower values.
    \item \textbf{Inner collection angle}: Making the inner collection angle smaller results in a clear trend of increased image intensity. With a smaller inner collection angle, more of the scattered electrons end up on the HAADF detector, which should make the image brighter. However, electrons scattered to different angles differ, and with a much smaller collection angle we would expect the contrast mechanism to change as more Bragg-diffracted electrons reach the detector. Such intricacies are likely not learned by the model, especially since the vast majority of images in the dataset were recorded with large collection angles relative to beam convergence angle.
\end{enumerate}

\FloatBarrier

\section{Pretraining Ablations}
\label{app:pretraining_ablations}

\paragraph{Backbone Sweep}
\label{app:backbone_sweep}

To select the image encoder, we compared four backbone configurations trained for 1000 epochs on the same train/val split (6{,}597/733 images) with a crop size of 256, effective batch size of 512, and metadata encoder $256 \times 3$. Table~\ref{tab:backbone_sweep} summarises the best-epoch retrieval metrics.

\begin{table}[H]
\centering
\caption{Backbone sweep for the CIMP image encoder (1000 epochs, crop size 256, effective batch 512, metadata encoder $256 \times 3$). ViT pretrained and ViT frozen use ImageNet-pretrained weights; the former is finetuned end-to-end while the latter keeps the backbone frozen and only trains the projection head.}
\label{tab:backbone_sweep}
\footnotesize
\setlength{\tabcolsep}{4pt}
\renewcommand{\arraystretch}{1.15}
\begin{tabular}{lrrr}
\toprule
\textbf{Backbone} & \textbf{Top-1} & \textbf{Top-5} & \textbf{Top-10} \\
\midrule
ResNet-18             & \textbf{0.844} & \textbf{0.969} & 0.984 \\
ViT (scratch)         & 0.750 & 0.953 & 0.984 \\
ViT (pretrained)      & 0.828 & \textbf{0.969} & \textbf{1.000} \\
ViT (pretrained, frozen) & 0.234 & 0.438 & 0.641 \\
\bottomrule
\end{tabular}
\end{table}

Perhaps surprisingly, ResNet-18 trained from scratch matches or exceeds the ImageNet-pretrained ViT (Top-1: 84.4\% vs.\ 82.8\%). This parity suggests that the distribution of HAADF-STEM micrographs differs substantially from natural images in ImageNet: the pretrained transformer features provide little transfer benefit for this domain, while a lightweight convolutional encoder with strong locality priors suffices to learn the relevant structure directly from the STEM data. Training a ViT from scratch lags both (Top-1: 75.0\%), as transformers tend to be more data-hungry when initialized randomly. Freezing the pretrained ViT and only training the projection head yields the worst results (Top-1: 23.4\%), which again suggests a large domain gap between STEM and natural images. Given this result, along with its smaller parameter count and simpler training pipeline, we adopt ResNet-18 as our default backbone for downstream style-transfer and denoising tasks.

\paragraph{Crop Size Sweep}

To determine the optimal image crop size for CIMP training, we swept over four crop sizes while holding the effective batch size fixed at $512$. All runs were trained for 1000 epochs on the same train/val split (6,597/733 images). Table~\ref{tab:crop_sweep} summarises the best-epoch retrieval metrics.

As shown in Table~\ref{tab:crop_sweep}, Crop 512 achieves the best Top-1 retrieval accuracy across the sweep, with smaller and larger crops both underperforming it. Crop 256 remains competitive on Top-5 and Top-10, while crop 1024 drops off noticeably, possibly because the reduced per-GPU batch and 8-way gradient accumulation weaken the per-step signal. We use crop 256 in subsequent experiments as it struck the balance between overall performance and practicalities for sensible batch sizes when training for downstream tasks. We observe that the retrieval accuracies across crops 128, 256, 512 are variable and somewhat similar. We take this as an indication that the acquisition parameters or "style" of an image is generally observable even at very small patches and relies relatively little on large-scale features. This varies somewhat by metadata parameter and is discussed in more depth in~\ref{sec:linear_probe}.

\begin{table}[H]
\centering
\caption{Crop size sweep for CIMP contrastive pre-training (1000 epochs each). ViT-pretrained backbone, metadata encoder $64 \times 2$, effective batch 512. For crop 1024, the per-GPU batch was reduced to 8 and gradients were accumulated over 8 steps to fit GPU memory.}
\label{tab:crop_sweep}
\footnotesize
\setlength{\tabcolsep}{4pt}
\renewcommand{\arraystretch}{1.15}
\begin{tabular}{rrrrrr}
\toprule
\textbf{Crop} & \textbf{Batch} & \textbf{Ep.} & \textbf{Top-1} & \textbf{Top-5} & \textbf{Top-10} \\
\midrule
128  & 512 & 1000 & 0.781 & 0.906 & 0.953 \\
256  & 512 & 1000 & 0.766 & \textbf{0.969} & \textbf{0.984} \\
512  & 512 & 1000 & \textbf{0.844} & 0.953 & 0.969 \\
1024 & 512 & 1000 & 0.738 & 0.913 & 0.950 \\
\bottomrule
\end{tabular}
\end{table}

\paragraph{Metadata Encoder Sweep}

We investigate the effect of varying the size of the metadata encoder. As shown in Table~\ref{tab:meta_sweep}, we find that scaling the metadata encoder beyond $64 \times 3$ yields modest returns: Top-1 improves from 0.797 at hidden dim 64 to 0.828 at 256, and saturates there. Top-10 reaches 1.000 at hidden dim 128 and above. We adopt $256 \times 3$ as the default metadata encoder configuration, which matches the Pareto frontier while keeping parameter count modest.

\begin{table}[H]
\centering
\caption{Metadata encoder capacity ablation. Best-model Top-$k$ retrieval accuracy. ViT-pretrained backbone, crop 256, effective batch 512, 1000 epochs. Width varied at depth 3.}
\label{tab:meta_sweep}
\footnotesize
\setlength{\tabcolsep}{3pt}
\renewcommand{\arraystretch}{1.15}
\begin{tabular}{rrrrr}
\toprule
\textbf{Hidden} & \textbf{Layers} & \textbf{Top-1} & \textbf{Top-5} & \textbf{Top-10} \\
\midrule
64   & 3 & 0.797 & 0.953 & 0.984 \\
128  & 3 & 0.766 & 0.953 & \textbf{1.000} \\
256  & 3 & \textbf{0.828} & 0.969 & \textbf{1.000} \\
512  & 3 & \textbf{0.828} & 0.969 & \textbf{1.000} \\
1024 & 3 & \textbf{0.828} & \textbf{0.984} & \textbf{1.000} \\
\bottomrule
\end{tabular}
\end{table}

\paragraph{Batch Size Sweep}

\begin{table}[H]
\centering
\caption{Effective batch size sweep. Best-model Top-$k$ retrieval accuracy. ResNet-18 backbone, crop 512, metadata encoder $256 \times 3$, 1000 epochs. As the validation set of our data contains only 733 samples, we do not explore batch sizes beyond 512. See Appendix~\ref{app:batch_sweep_vit} for the corresponding ViT-pretrained sweep.}
\label{tab:batch_sweep}
\footnotesize
\setlength{\tabcolsep}{4pt}
\renewcommand{\arraystretch}{1.15}
\begin{tabular}{rrrr}
\toprule
\textbf{Eff.\ batch} & \textbf{Top-1} & \textbf{Top-5} & \textbf{Top-10} \\
\midrule
128 & 0.775 & 0.988 & \textbf{1.000} \\
256 & \textbf{0.859} & \textbf{1.000} & \textbf{1.000} \\
512 & \textbf{0.859} & \textbf{1.000} & \textbf{1.000} \\
\bottomrule
\end{tabular}
\end{table}

As shown in Table~\ref{tab:batch_sweep}, retrieval accuracy improves noticeably from effective batch 128 (Top-1: 0.775) to 256 (Top-1: 0.859) and saturates thereafter. Since our validation set contains only 733 samples, batch sizes beyond 512 would exceed reasonable negative-to-positive ratios; we use 512 as the default.

\label{app:batch_sweep_vit}
For completeness we reproduce the effective batch size sweep with the ViT-pretrained backbone, the performance of which is shown in Table~\ref{tab:batch_sweep_vit}.

\begin{table}[H]
\centering
\caption{Effective batch size sweep with ViT-pretrained backbone. Best-model Top-$k$ retrieval accuracy. Crop 256, metadata encoder $256 \times 3$, 1000 epochs.}
\label{tab:batch_sweep_vit}
\footnotesize
\setlength{\tabcolsep}{4pt}
\renewcommand{\arraystretch}{1.15}
\begin{tabular}{rrrr}
\toprule
\textbf{Eff.\ batch} & \textbf{Top-1} & \textbf{Top-5} & \textbf{Top-10} \\
\midrule
128 & 0.775 & 0.950 & 0.975 \\
256 & \textbf{0.828} & \textbf{0.969} & \textbf{1.000} \\
512 & \textbf{0.828} & \textbf{0.969} & \textbf{1.000} \\
\bottomrule
\end{tabular}
\end{table}

\section{CIMP Embedding Visualization}
\label{app:cimp_embedding_visualization}

\begin{figure}[H]
    \centering
    \includegraphics[width=0.7\linewidth]{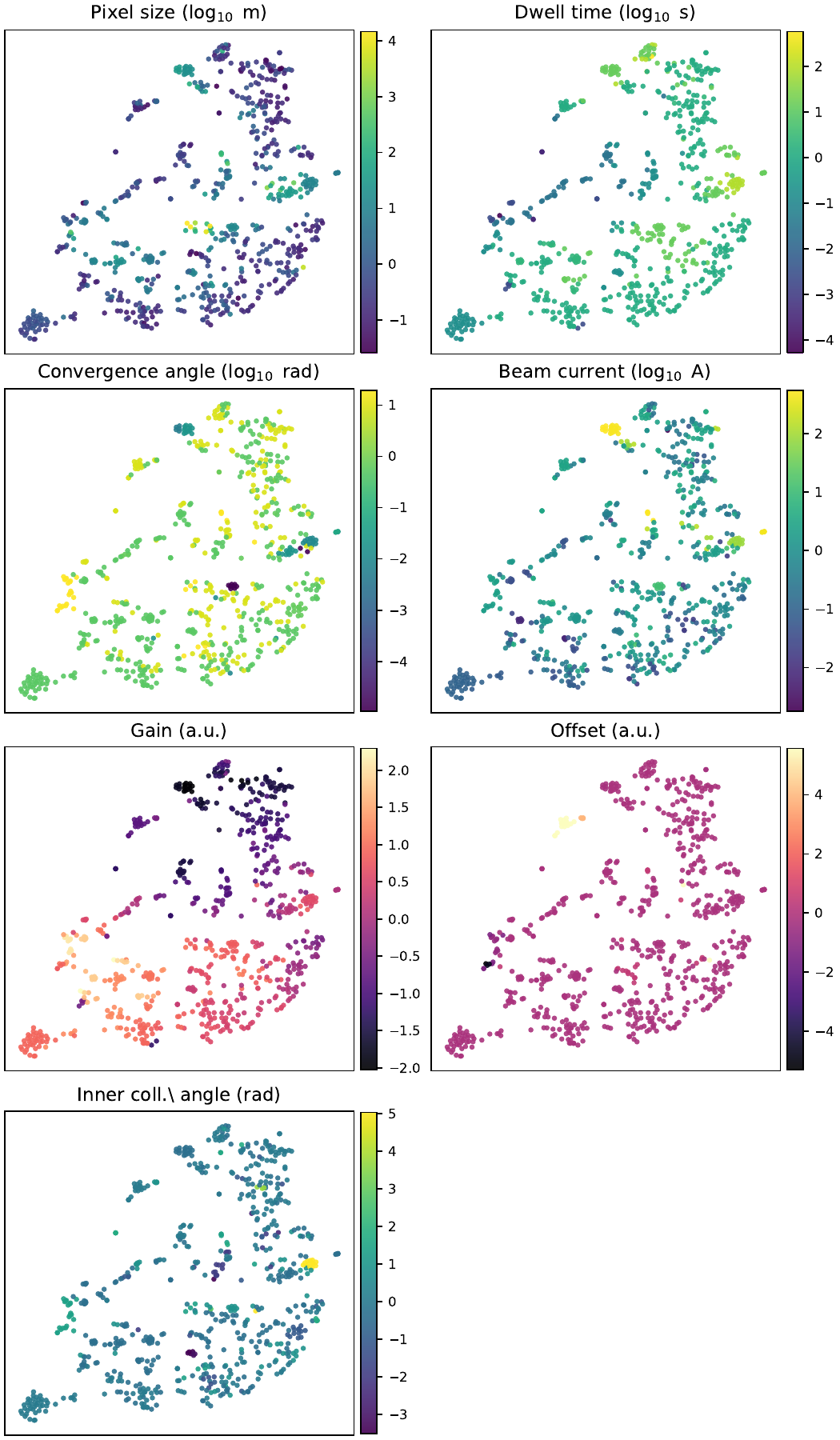}
    \caption{t-SNE of the frozen CIMP visual embeddings on the held-out validation set ($n=733$), with the same 2D layout coloured by each of the seven metadata dimensions in turn. Gain shows the cleanest monotonic gradient and pixel size shows clear viridis bands, while the discrete-valued dimensions (convergence angle, inner collection angle) form pockets aligned with their few populated values. All metadata values are z-scored. Because the visual embedding also encodes image content, the layout reflects content variation in addition to metadata, and any apparent metadata gradient should be read as one factor among several driving the embedding geometry and its 2D projection.}
    \label{fig:tsne_metadata_panels}
\end{figure}

We project the frozen CIMP visual embeddings on the held-out validation set ($n=733$) to two dimensions with t-SNE~\citep{vandermaaten2008tsne} (perplexity $30$, cosine metric, PCA initialisation, fixed seed). The same 2D layout is reused across both figures in this section. Figure~\ref{fig:tsne_metadata_panels} colours the layout by each of the seven metadata dimensions in turn. 

\section{Style Transfer Embedding Verification}
\label{app:style_transfer_embedding_verification}

\begin{figure}[H]
    \centering
    \includegraphics[width=0.65\linewidth]{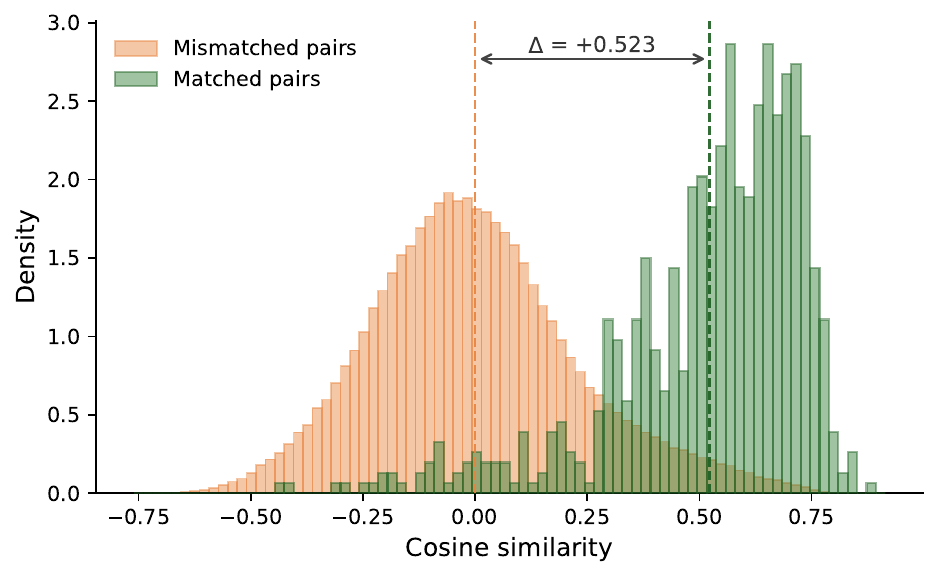}
    \caption{Cosine similarity distribution for style transfer. Style-transferred validation images are passed through the CIMP visual encoder, and the resulting embeddings are compared to their target metadata embedding (correct pairs) and to all other metadata embeddings in the validation batch (incorrect pairs). The mean similarity of correct pairs exceeds that of incorrect pairs by $0.523$, indicating that the generator aligns image style with the provided conditioning embedding rather than producing arbitrary outputs.}
    \label{fig:sim_dist_gen_images}
\end{figure}

\section{GAN Ablations}
\label{app:gan_ablations}

We swept loss weights, the cycle-consistency masking ratio, and a noise-augmentation toggle for the metadata-conditioned style-transfer GAN. All runs share the same CIMP encoder, U-Net architecture, and optimiser, and were trained for 100 epochs. Table~\ref{tab:gan_ablations} reports validation losses at the best composite-validation epoch within each run.

\begin{table}[H]
\centering
\caption{GAN ablations, each run trained for 100 epochs. Validation embedding-alignment, identity, and cycle losses, and the best composite total validation loss within each run.}
\label{tab:gan_ablations}
\footnotesize
\setlength{\tabcolsep}{3pt}
\renewcommand{\arraystretch}{1.15}
\begin{tabular}{rrrrrrrrrrr}
\toprule
\textbf{Run} & $\lambda_{\text{emb}}$ & $\lambda_{\text{GAN}}$ & $\lambda_{\text{cyc}}$ & $\lambda_{\text{id}}$ & \textbf{mask} & \textbf{noise} & val\_emb $\downarrow$ & val\_id $\downarrow$ & val\_cyc $\downarrow$ & best total\_val $\downarrow$ \\
\midrule
 1 & 1.0 & 1.0 & 0.1   & 0.1   &  5\% & off & 0.00801 & 0.01337 & 0.00993 & 0.01005 \\
 2 & 1.0 & 1.0 & 0.2   & 0.1   &  5\% & off & 0.00906 & 0.01456 & 0.01619 & 0.01364 \\
 3 & 1.0 & 1.0 & 0.05  & 0.1   &  5\% & off & 0.00926 & 0.00321 & 0.00745 & 0.00978 \\
 4 & 1.0 & 1.0 & 0.1   & 0.05  &  5\% & off & 0.00778 & 0.01293 & 0.01103 & 0.00954 \\
 5 & 1.0 & 1.0 & 0.1   & 0.1   &  5\% & on  & 0.00824 & 0.00277 & 0.00653 & 0.00904 \\
 6 & 1.0 & 1.0 & 0.05  & 0.05  &  5\% & off & 0.01011 & 0.00465 & 0.01137 & 0.01093 \\
 7 & 1.0 & 1.0 & 0.1   & 0.075 &  0\% & off & 0.00944 & 0.00321 & 0.00641 & 0.01017 \\
 8 & 1.0 & 1.0 & 0.1   & 0.1   & 15\% & off & 0.00759 & 0.00364 & 0.00813 & 0.00852 \\
 9 & 1.0 & 1.0 & 0.1   & 0.1   & 10\% & off & 0.00768 & 0.00281 & 0.00747 & 0.00859 \\
10 & 2.0 & 1.0 & 0.1   & 0.1   &  5\% & off & 0.00768 & 0.01377 & 0.01902 & 0.01864 \\
11 & 1.0 & 0   & 0.1   & 0.1   &  5\% & off & 0.00702 & 0.01921 & 0.00835 & 0.00978 \\
\bottomrule
\end{tabular}
\end{table}

Several runs achieved lower validation losses than Run 1, but qualitative inspection of their generated images revealed visually bizarre outputs that were inconsistent with the target metadata and input content, so we selected Run 1 as the best-performing setting. When we extended Run 1 to 250 epochs, validation performance degraded, and we increased $\lambda_{\text{cyc}}$ from 0.1 to 0.3 to counteract that drift. The final configuration reported in the main text was trained for 250 epochs with $\lambda_{\text{emb}} = \lambda_{\text{GAN}} = 1$, $\lambda_{\text{cyc}} = 0.3$, and $\lambda_{\text{id}} = 0.1$.

\paragraph{Training Histories}
\label{app:training_histories}

\begin{figure}[H]
    \centering
    \includegraphics[width=\linewidth]{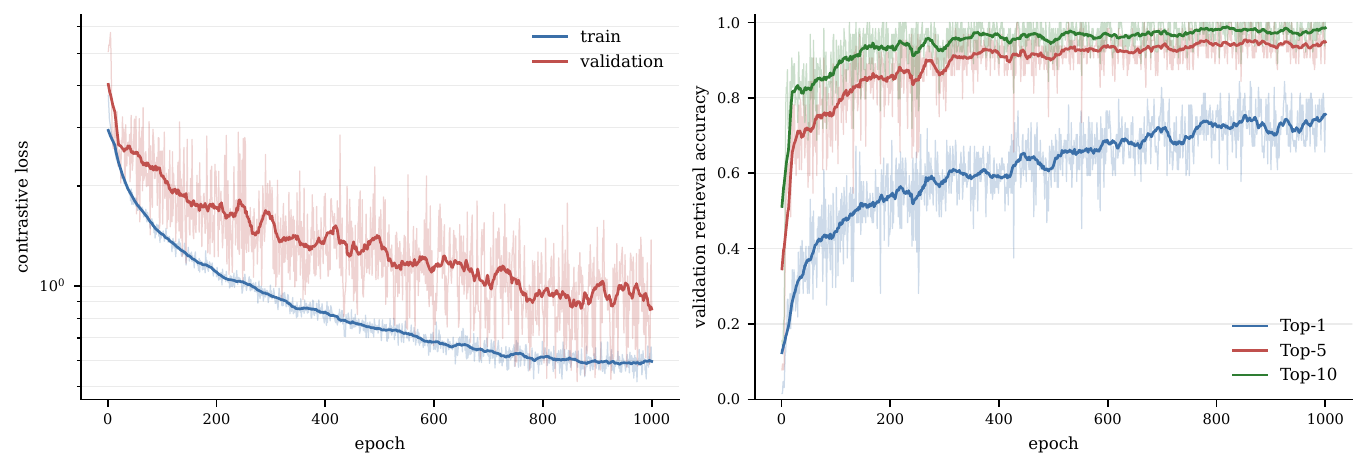}
    \caption{Training history of the best CIMP encoder. Left: train and validation contrastive loss (log scale). Right: validation retrieval accuracy (Top-1, Top-5, Top-10).}
    \label{fig:training_history_cimp}
\end{figure}

\begin{figure}[H]
    \centering
    \includegraphics[width=\linewidth]{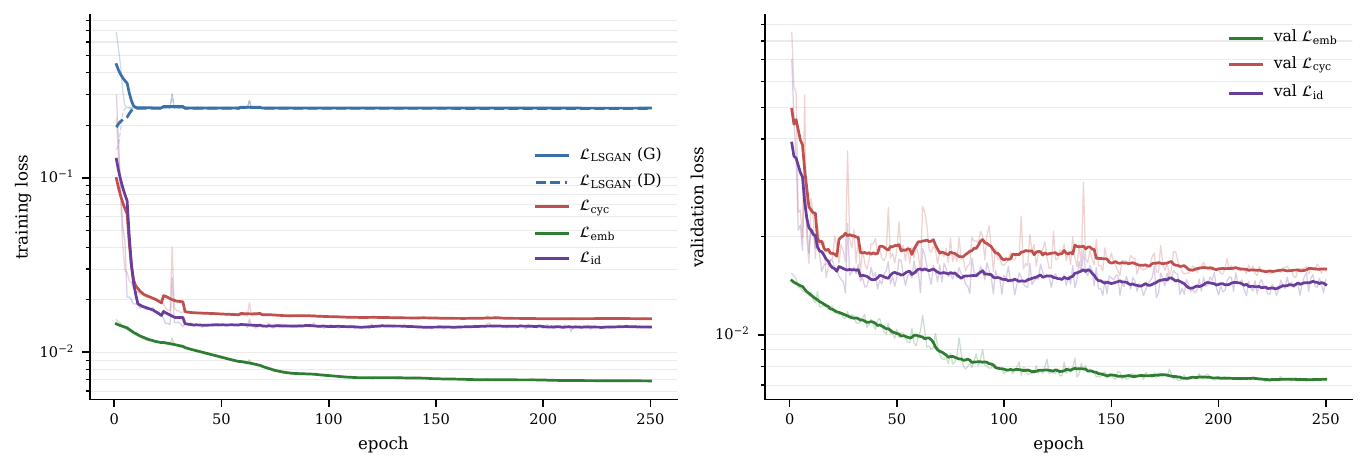}
    \caption{Training history of the style-transfer GAN. Left: training losses per term, including the LSGAN generator / discriminator pair and the three non-adversarial components. Right: validation losses for the three non-adversarial components.}
    \label{fig:training_history_gan}
\end{figure}

Figures~\ref{fig:training_history_cimp} and~\ref{fig:training_history_gan} show the per-epoch training histories for the two learned models reported in the main text: the best CIMP encoder (ResNet-18, crop 256, metadata encoder $256 \times 3$, effective batch 512, 1000 epochs) and the metadata-conditioned style-transfer GAN trained on top of it (StyleUNet with FiLM conditioning, base filters 32, crop 256, batch 8, 250 epochs). Bold curves show a rolling mean (25 epochs for CIMP, 10 epochs for the GAN); raw per-epoch values are overlaid in faint.

\section{Denoiser Metrics and Power Spectra}
\label{app:denoiser_psd}

Table~\ref{tab:denoise_metrics_full} reports the full set of reference-free denoising metrics for the 100 held-out validation frames analyzed in the main text. The two metrics omitted from the main-text table, residual std and the SNR proxy, are included for completeness. Both reward conservative behavior. A denoiser that returned the input unchanged would score perfectly on residual std and would have an unbounded SNR proxy, so we treat them as context rather than as primary measures of denoising quality.

\begin{table}[H]
\centering
\caption{Full reference-free denoising metric set on 100 held-out validation frames. Values are mean $\pm$ standard error of the mean across frames. Residual std measures how much the denoiser changed the input. The SNR proxy is the variance ratio of denoised output to residual expressed in dB. Higher is better for gradient correlation and edge energy ratio. The bottom three rows report $\log_{10}$ ratios of raw-to-denoised radial power averaged within each frequency band, where positive values indicate suppression and negative values indicate added power.}
\label{tab:denoise_metrics_full}
\footnotesize
\begin{tabular}{lccc}
\toprule
Metric & CIMP GAN (ours) & Noise2Void & Noise2Atom \\
\midrule
Residual std                                    & $0.023 \pm 0.002$  & $0.015 \pm 0.002$    & $0.060 \pm 0.005$   \\
Residual mean                                   & $0.007 \pm 0.002$  & $0.0001 \pm 0.00005$ & $-0.088 \pm 0.010$  \\
Gradient correlation~$\uparrow$                 & $0.57 \pm 0.02$    & $0.69 \pm 0.01$      & $0.24 \pm 0.01$     \\
Edge energy ratio~$\uparrow$                    & $0.79 \pm 0.03$    & $0.36 \pm 0.01$      & $0.58 \pm 0.07$     \\
SNR proxy (dB)                                  & $4.9 \pm 0.7$      & $10.1 \pm 0.6$       & $-2.9 \pm 0.3$      \\
\midrule
PSD ratio, low-freq ($\log_{10}$)                & $0.18 \pm 0.03$    & $-0.012 \pm 0.001$   & $0.69 \pm 0.06$     \\
PSD ratio, mid-freq ($\log_{10}$)                & $0.24 \pm 0.03$    & $0.44 \pm 0.02$      & $-0.09 \pm 0.03$    \\
PSD ratio, high-freq ($\log_{10}$)~$\uparrow$    & $0.25 \pm 0.03$    & $1.08 \pm 0.02$      & $0.87 \pm 0.07$     \\
\bottomrule
\end{tabular}
\end{table}

\begin{figure}[H]
    \centering
    \includegraphics[width=\linewidth]{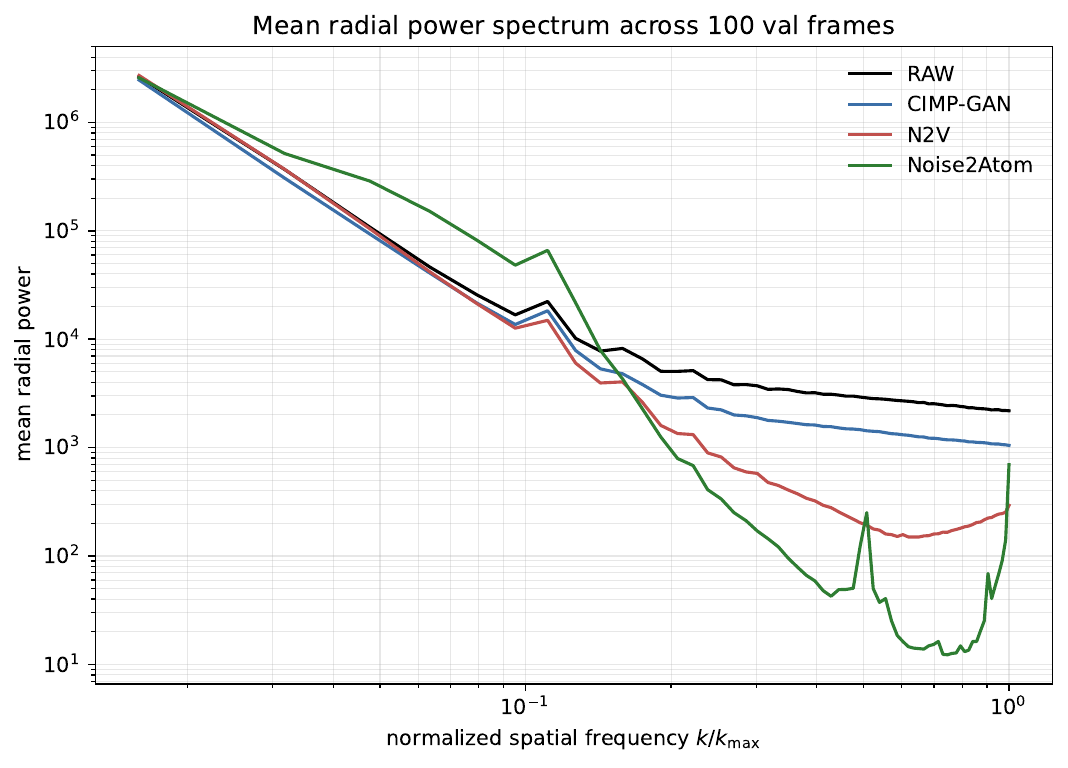}
    \caption{Mean radial power spectrum across 100 held-out validation frames for the raw input and each denoiser, plotted on log-log axes against normalized spatial frequency $k/k_{\max}$. Noise2Void aggressively suppresses high-frequency power. CIMP GAN suppresses power gently and uniformly. Noise2Atom adds power at low-to-mid frequencies before crashing at high frequencies, with prominent peaks at the spatial frequencies of its learned atomic prior.}
    \label{fig:psd}
\end{figure}

Figure~\ref{fig:psd} shows the full mean radial power spectrum of each denoiser averaged over the same 100 frames. The three frequency-band $\log_{10}$ ratios reported in Table~\ref{tab:denoise_metrics_full} are computed by averaging these curves within their respective bands.

\section{Noise2Atom Sample Outputs}
\label{app:n2a_samples}

\begin{figure}[H]
    \centering
    \begin{subfigure}{\linewidth}
        \centering
        \begin{minipage}{0.32\linewidth}\centering
            \footnotesize\textbf{Raw}\\[2pt]
            \includegraphics[width=\linewidth]{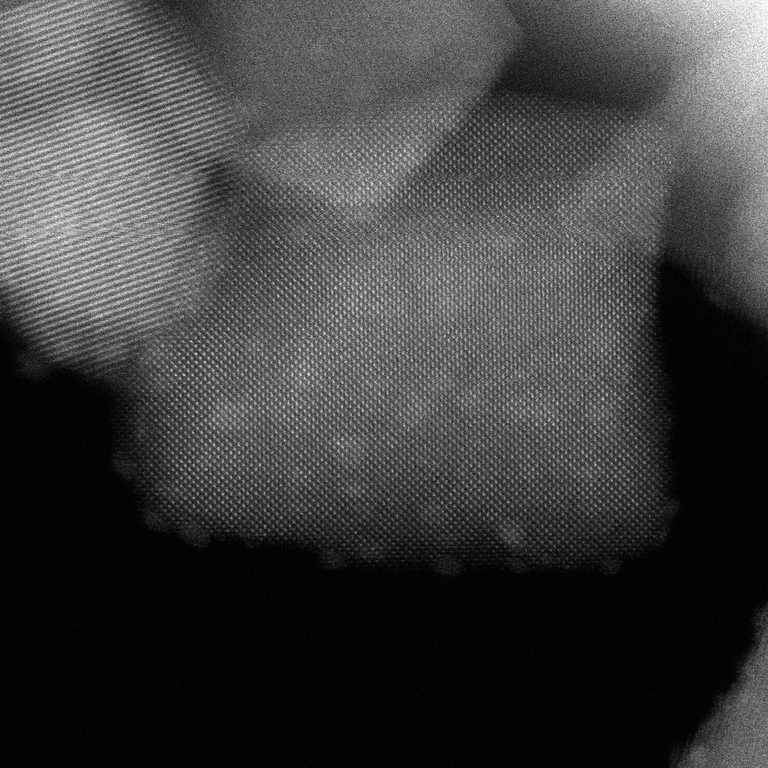}
        \end{minipage}\hfill
        \begin{minipage}{0.32\linewidth}\centering
            \footnotesize\textbf{Noise2Atom}\\[2pt]
            \includegraphics[width=\linewidth]{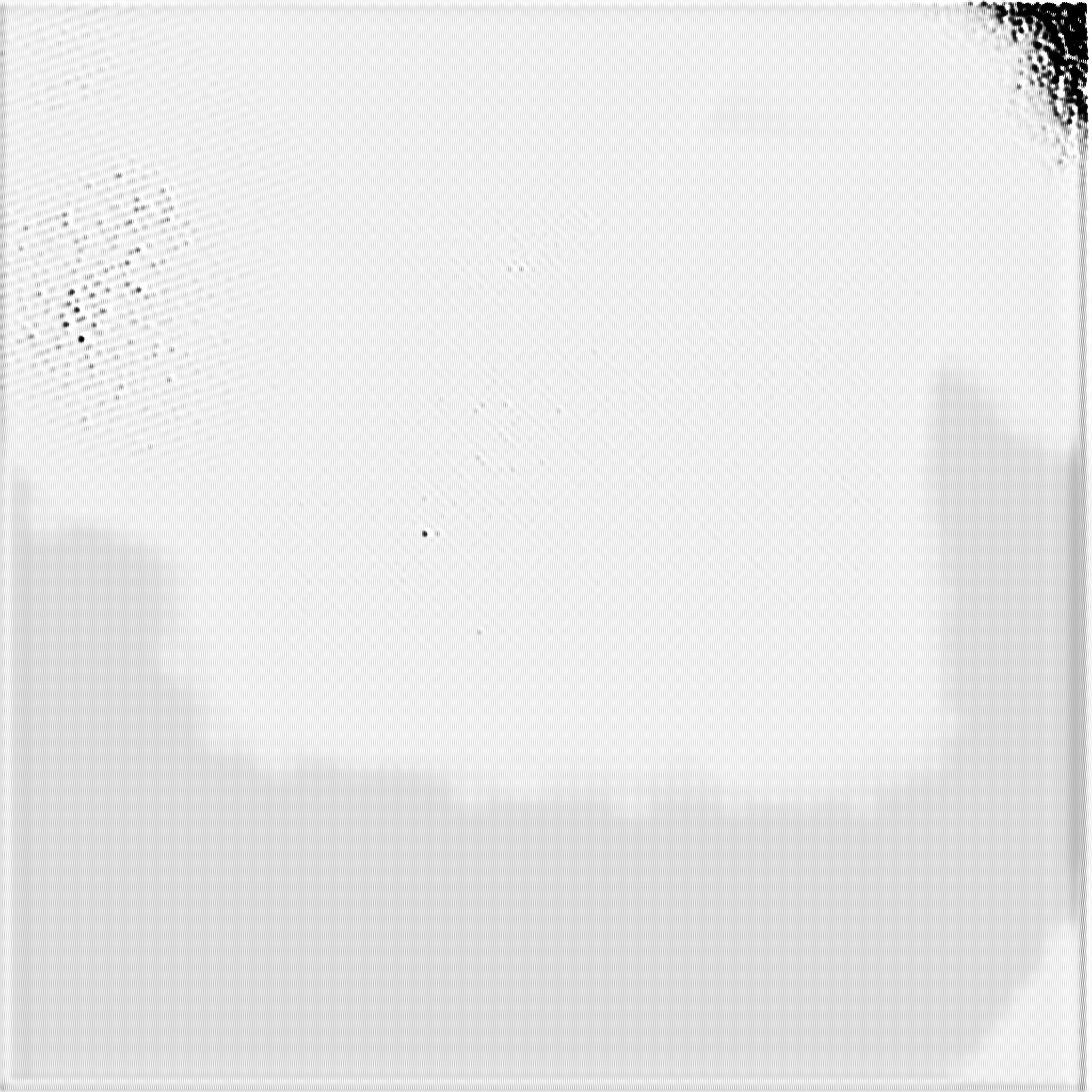}
        \end{minipage}\hfill
        \begin{minipage}{0.32\linewidth}\centering
            \footnotesize\textbf{Residual}\\[2pt]
            \includegraphics[width=\linewidth]{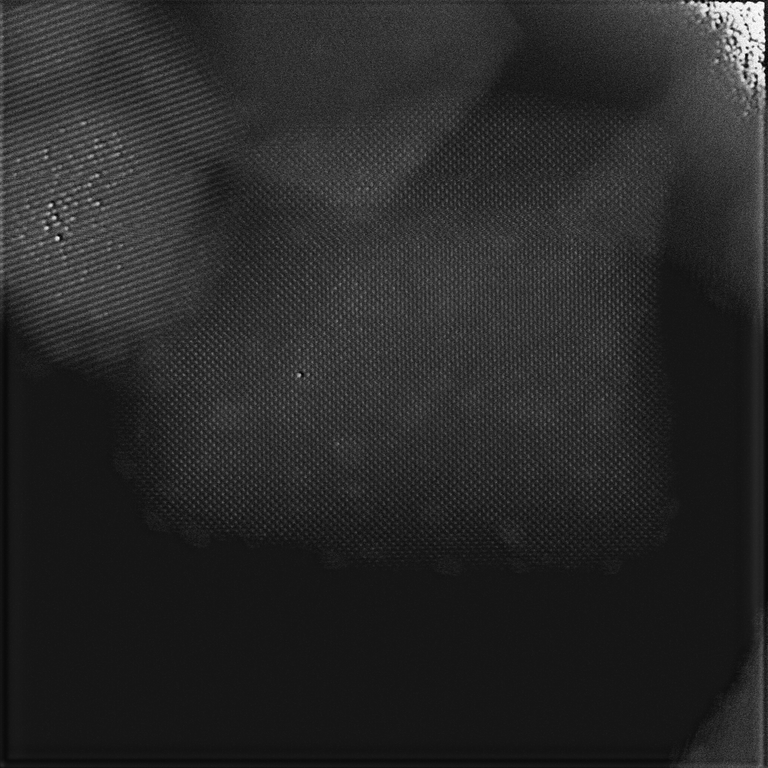}
        \end{minipage}
        \caption{\ce{Pt/CeO2} small particle (idx 3716).}
        \label{fig:n2a_a}
    \end{subfigure}

    \vspace{4pt}
    \begin{subfigure}{\linewidth}
        \centering
        \begin{minipage}{0.32\linewidth}\centering
            \includegraphics[width=\linewidth]{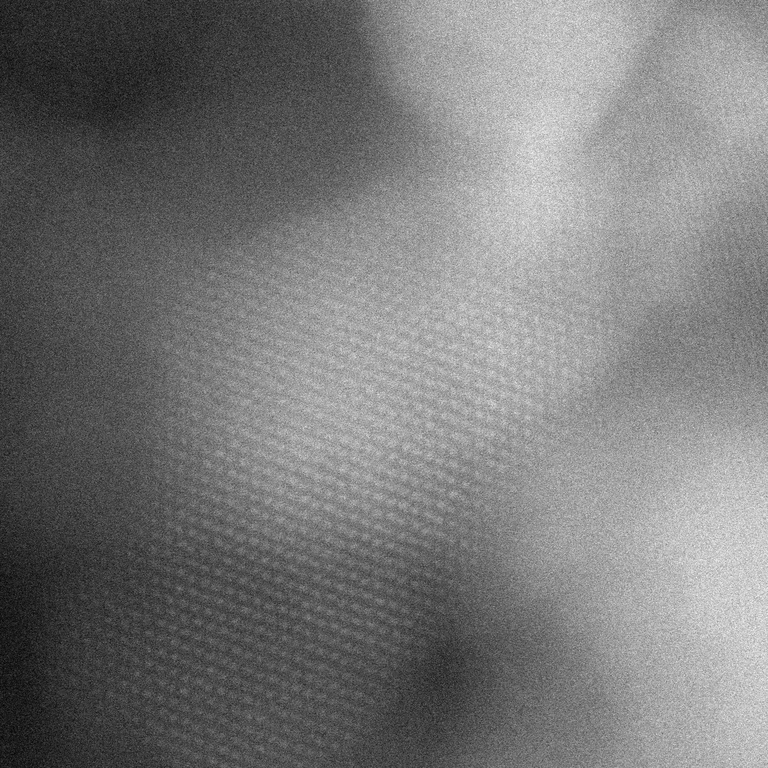}
        \end{minipage}\hfill
        \begin{minipage}{0.32\linewidth}\centering
            \includegraphics[width=\linewidth]{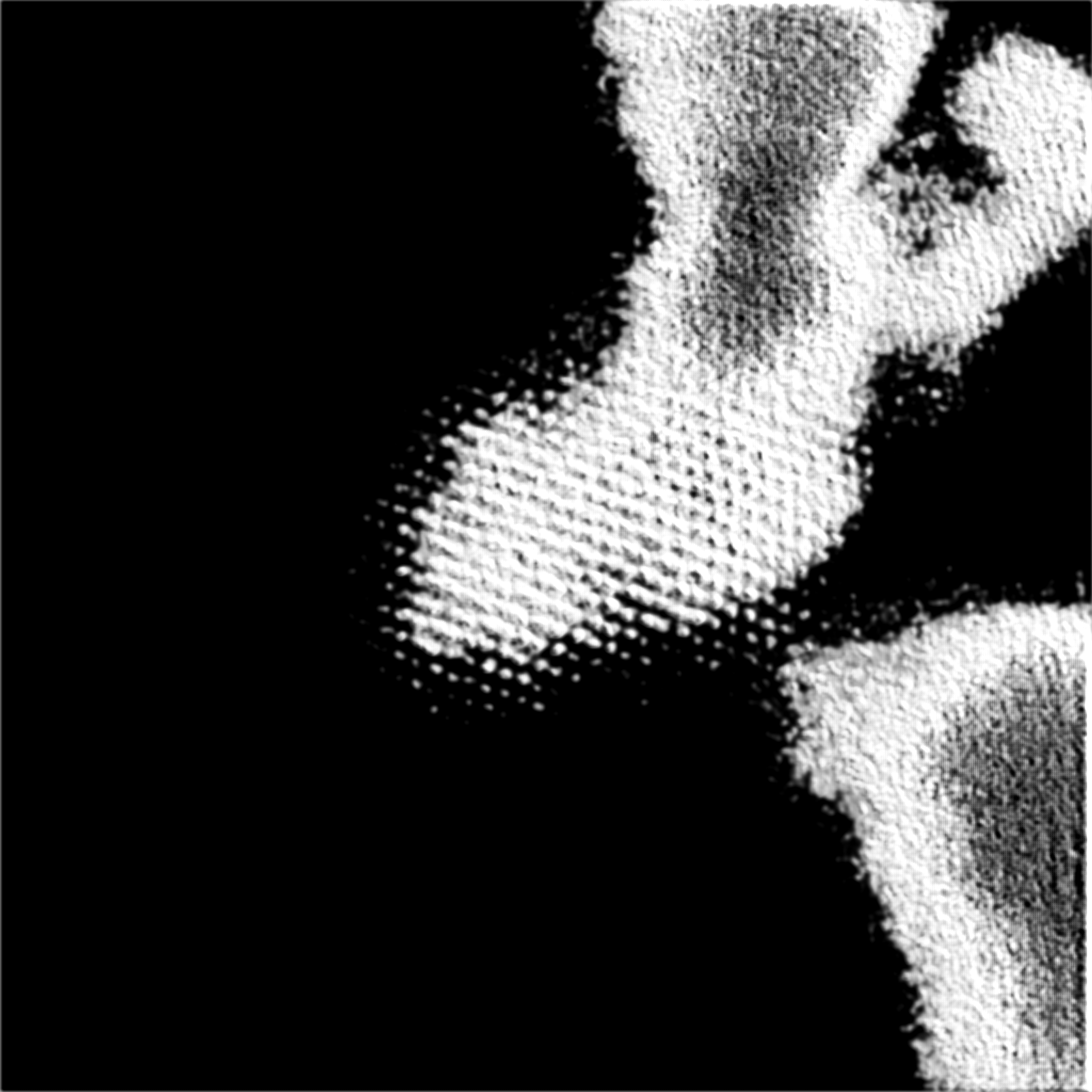}
        \end{minipage}\hfill
        \begin{minipage}{0.32\linewidth}\centering
            \includegraphics[width=\linewidth]{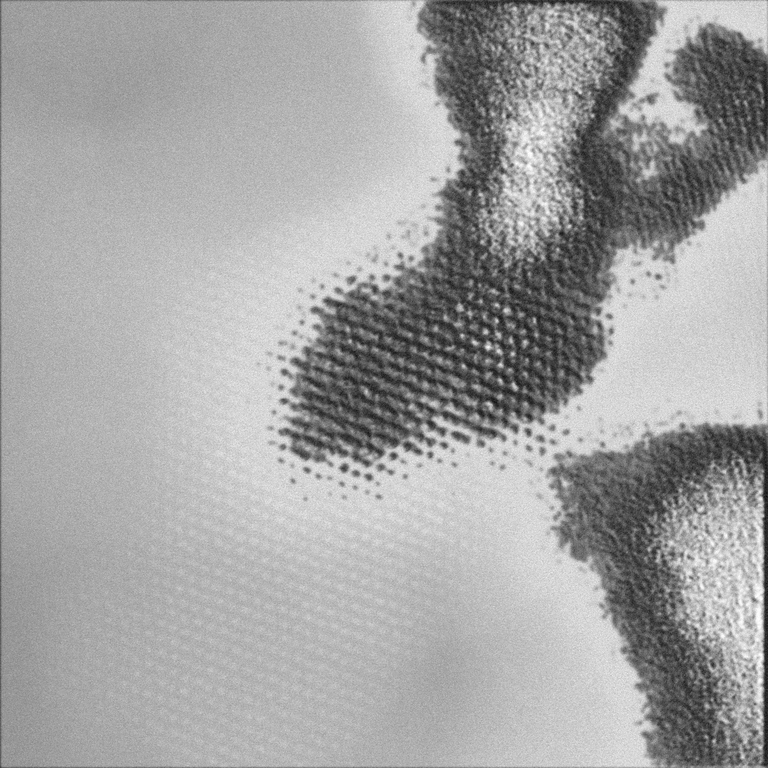}
        \end{minipage}
        \caption{\ce{Pd/In/ZrO2} (in-situ), atomic resolution (idx 4969).}
        \label{fig:n2a_b}
    \end{subfigure}

    \vspace{4pt}
    \begin{subfigure}{\linewidth}
        \centering
        \begin{minipage}{0.32\linewidth}\centering
            \includegraphics[width=\linewidth]{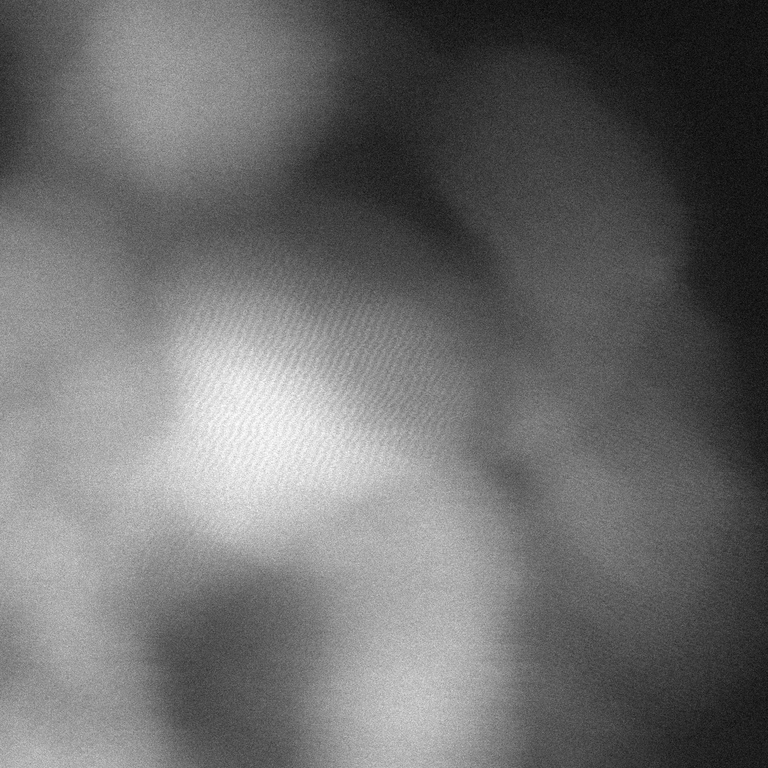}
        \end{minipage}\hfill
        \begin{minipage}{0.32\linewidth}\centering
            \includegraphics[width=\linewidth]{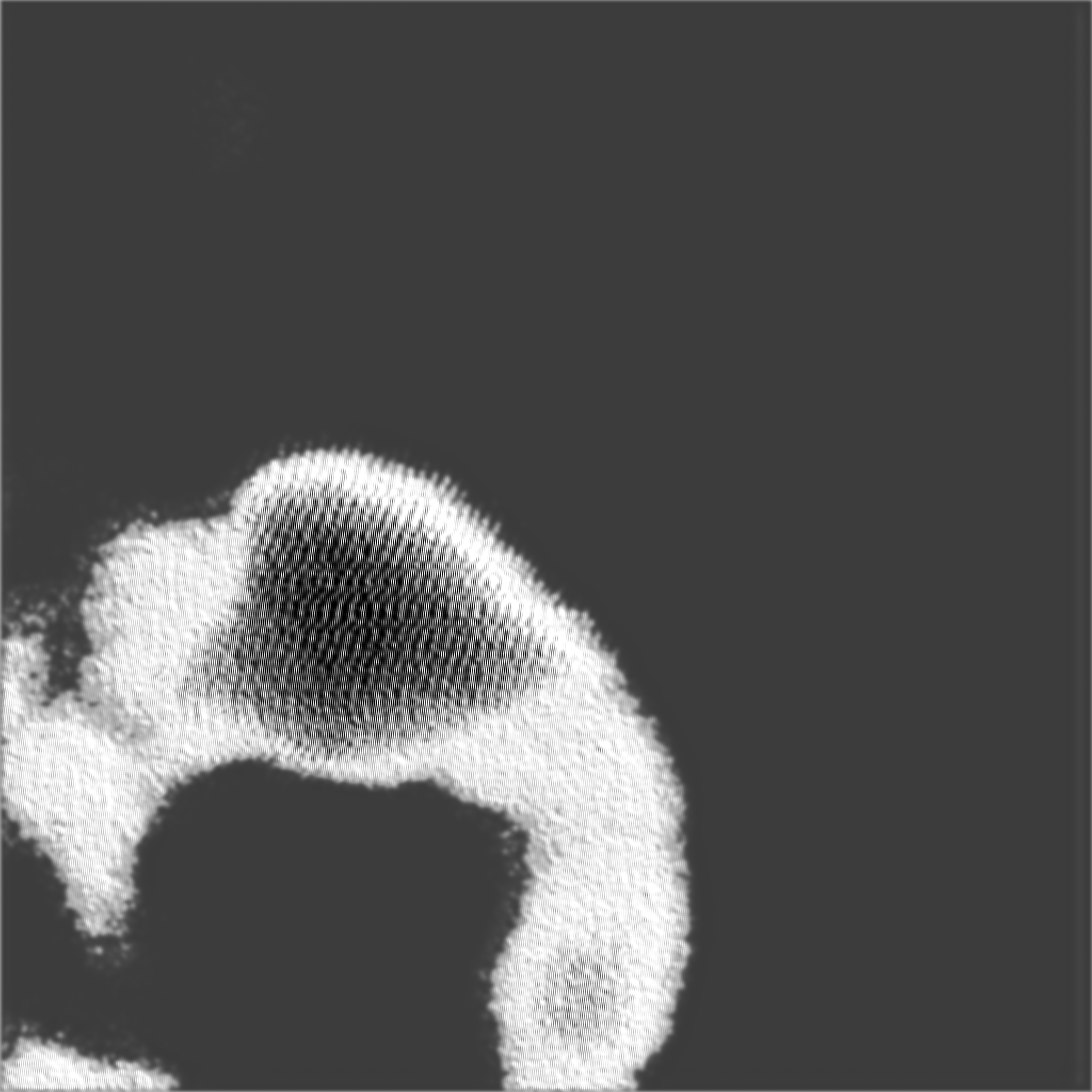}
        \end{minipage}\hfill
        \begin{minipage}{0.32\linewidth}\centering
            \includegraphics[width=\linewidth]{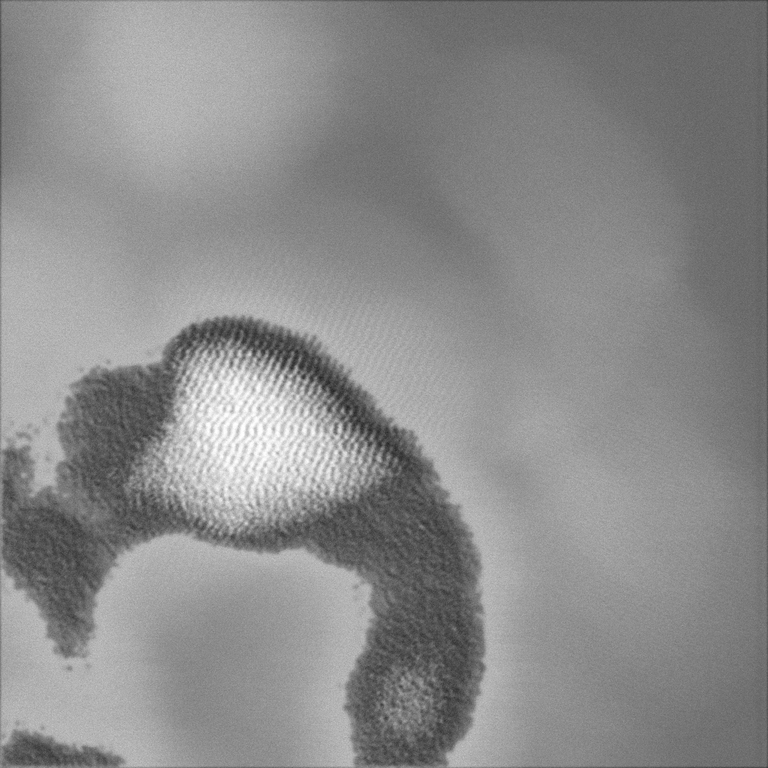}
        \end{minipage}
        \caption{\ce{Pd/In/ZrO2} (in-situ), sub-atomic field (idx 4656).}
        \label{fig:n2a_c}
    \end{subfigure}

    \vspace{4pt}
    \begin{subfigure}{\linewidth}
        \centering
        \begin{minipage}{0.32\linewidth}\centering
            \includegraphics[width=\linewidth]{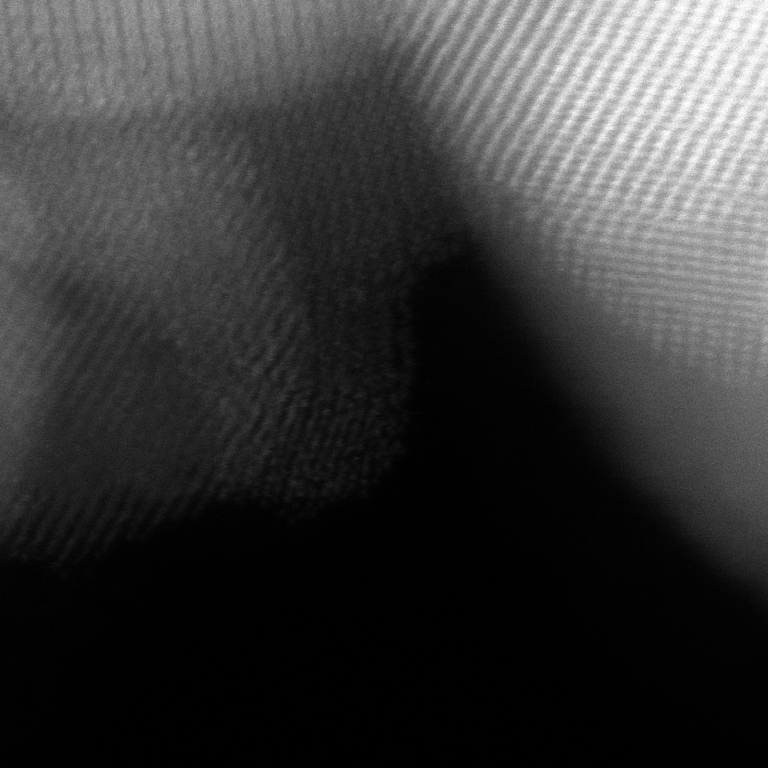}
        \end{minipage}\hfill
        \begin{minipage}{0.32\linewidth}\centering
            \includegraphics[width=\linewidth]{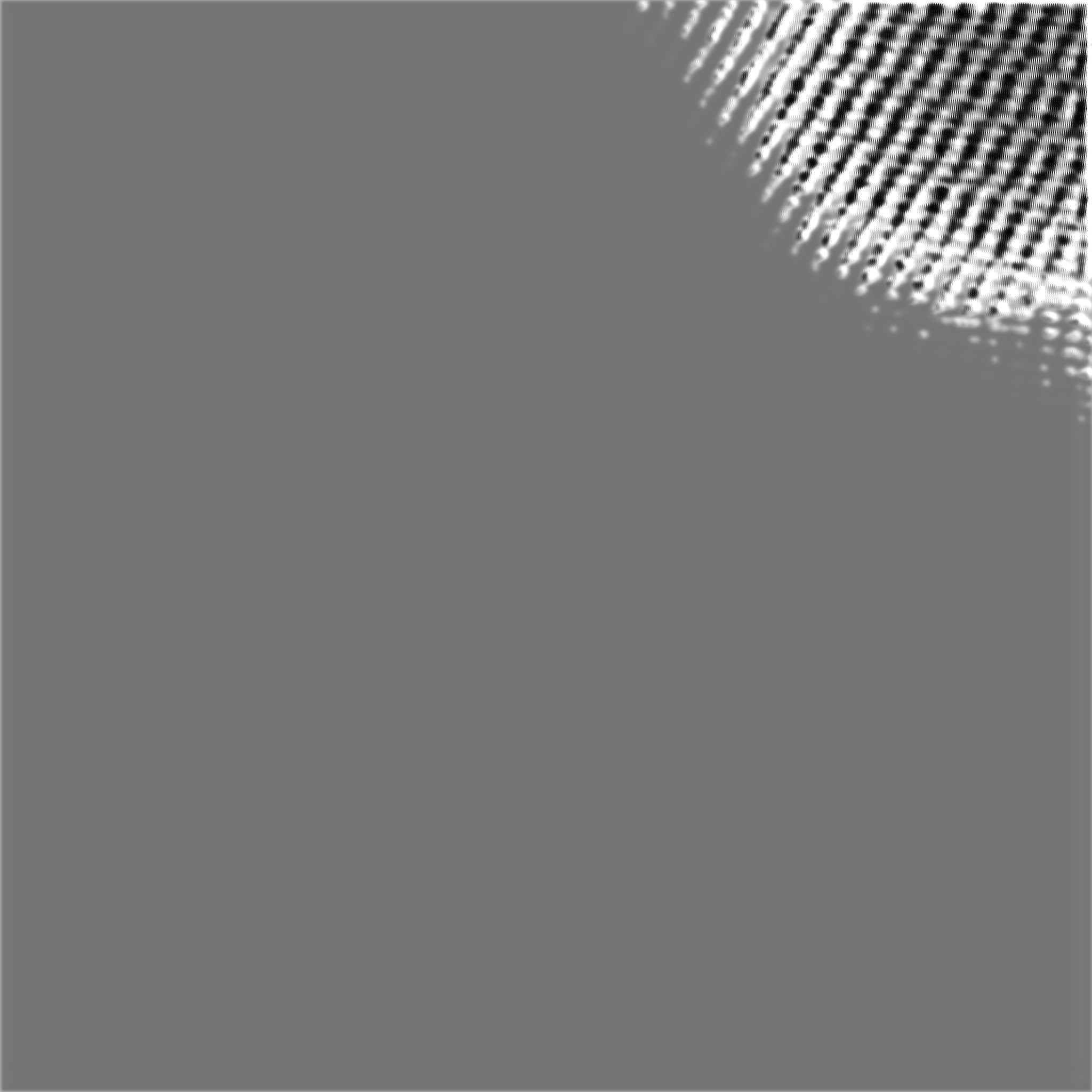}
        \end{minipage}\hfill
        \begin{minipage}{0.32\linewidth}\centering
            \includegraphics[width=\linewidth]{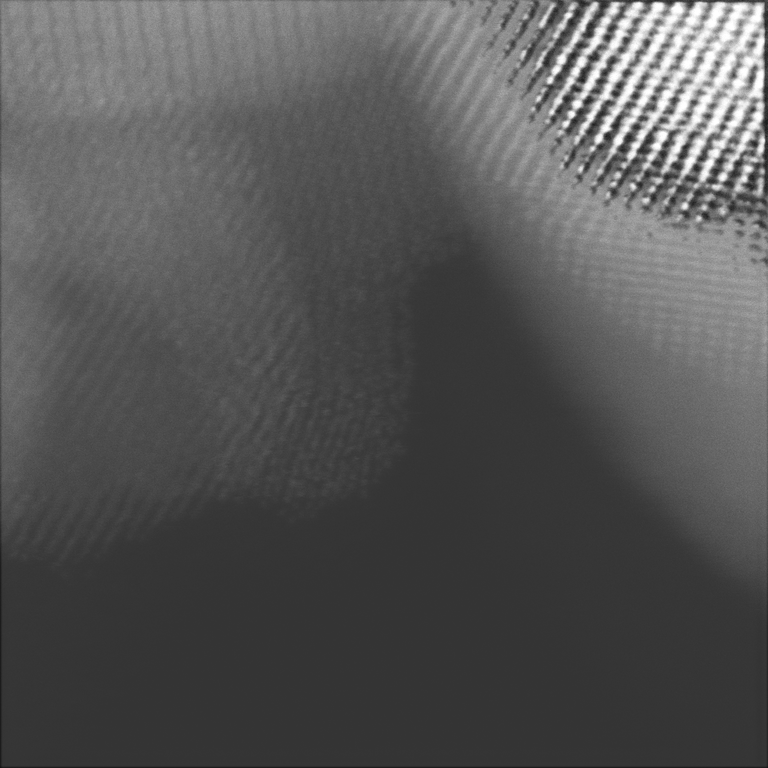}
        \end{minipage}
        \caption{$2\,\text{wt}\%$ \ce{In2O3/HfO2} from flame-spray pyrolysis (idx 6250).}
        \label{fig:n2a_d}
    \end{subfigure}

    \caption{Four held-out frames from the validation set denoised by Noise2Atom. Within each row, columns are raw acquisition, Noise2Atom output, and residual (raw minus denoised). We observe that the model only works on clean atom-lattice sections of the images and otherwise imprints periodic structure on non-atomic content (\subref{fig:n2a_a}, \subref{fig:n2a_c}) or collapses the image into a near-uniform shade while keeping only the brightest features (\subref{fig:n2a_d}).}
    \label{fig:n2a_samples}
\end{figure}

Figure~\ref{fig:n2a_samples} shows four representative held-out frames denoised by Noise2Atom, drawn from the same 100-frame validation set used for the metric computation in Table~\ref{tab:denoise_metrics_full}. Each row shows the raw acquisition, the Noise2Atom output, and the residual (raw minus denoised). The four examples span the range of behaviours described in Section~\ref{sec:denoising} and confirmed by the metrics. The atomic-resolution \ce{Pd/In/ZrO2} frame in panel~\subref{fig:n2a_b} is the regime Noise2Atom was designed for, and its output recovers a clean lattice of bright atomic peaks. The remaining three frames sit outside this regime and show the failure modes named in the main text. In the \ce{Pt/CeO2} small-particle frame in panel~\subref{fig:n2a_a}, the model imprints periodic structure on regions where no atomic resolution is present in the input. The \ce{In2O3/HfO2} FSP frame in panel~\subref{fig:n2a_d} shows the most extreme case. Here, the model collapses most of the image to a near-uniform colour while preserving only the brightest corner. The non-zero residuals in every panel reflect the systematic intensity shift captured by the negative residual mean of $-0.088$ in Table~\ref{tab:denoise_metrics_full}.

\section{User Study}
\label{app:user_study}

Five experimental microscopists submitted judgments, with uneven coverage across the $200$ test triplets. The raters were not compensated in any way. Each rater independently chose between the Noise2Void and CIMP GAN versions of every triplet they were shown, with left/right placement randomised per triplet. Their per-rater preference is reported in Table~\ref{tab:user_study_prelim} of the main text, with Wilson $95\%$ confidence intervals on each rate and a binomial $p$-value against the $50/50$ null. Here we focus on overlap and rater agreement, with corresponding confidence intervals and $p$-values.

\paragraph{Instructions to raters.} Each rater was shown the following prompt at the start of the study:
\begin{displayquote}
On each screen you'll see a raw noisy image at the top and two denoised versions of it below, labelled A and B. The two versions come from different denoising approaches, presented in a random left/right order so the comparison stays blind. Hold Shift and scroll over an image to zoom (up to $10\times$), drag to pan, double-click to reset.
\end{displayquote}

\paragraph{Coverage.} Every triplet has received at least one judgment. The distribution of raters per triplet is given in Table~\ref{tab:user_study_coverage}.

\begin{table}[h]
\centering
\caption{Number of triplets covered by exactly $k$ raters.}
\label{tab:user_study_coverage}
\begin{tabular}{lr}
\toprule
Raters per triplet & Triplets \\
\midrule
1 & 29 \\
2 & 79 \\
3 & 70 \\
4 & 21 \\
5 &  1 \\
\midrule
Total & 200 \\
\bottomrule
\end{tabular}
\end{table}

\paragraph{Pairwise agreement.} For triplets judged by both members of a pair, the agreement rate is the fraction of overlapping triplets where the two raters chose the same denoiser. Cohen's $\kappa$ corrects for chance agreement implied by each rater's marginal preference, so a $\kappa$ near $0$ indicates that the agreement rate is no better than would be expected from the marginals alone. We report $\kappa$ with its asymptotic standard error and a one-sided $p$-value against the null $\kappa = 0$.

\begin{table}[h]
\centering
\caption{Pairwise agreement on overlapping triplets, with Wilson $95\%$ CI on agreement and Cohen's $\kappa$ with its asymptotic standard error. The $p$-value is one-sided against $\kappa = 0$.}
\label{tab:user_study_overlap}
\footnotesize
\begin{tabular}{lrlll}
\toprule
Pair & Overlap & Agreement (Wilson $95\%$ CI) & Cohen's $\kappa$ ($\pm$SE) & $p$ vs $\kappa{=}0$ \\
\midrule
Rater 1, Rater 2 & 101 & 77 (76.2\% [67.1, 83.5]) & $\phantom{-}0.43 \pm 0.10$ & $0.01$ \\
Rater 1, Rater 3 &  76 & 59 (77.6\% [67.1, 85.5]) & $\phantom{-}0.33 \pm 0.14$ & $0.13$ \\
Rater 1, Rater 4 &  73 & 39 (53.4\% [42.1, 64.4]) & $\phantom{-}0.06 \pm 0.12$ & $0.36$ \\
Rater 1, Rater 5 &  36 & 22 (61.1\% [44.9, 75.2]) & $\phantom{-}0.17 \pm 0.17$ & $0.28$ \\
Rater 2, Rater 3 &  40 & 32 (80.0\% [65.2, 89.5]) & $\phantom{-}0.38 \pm 0.19$ & $0.18$ \\
Rater 2, Rater 4 &  33 & 18 (54.5\% [38.0, 70.2]) &           $-0.05 \pm 0.20$ & $0.56$ \\
Rater 2, Rater 5 &  14 &  9 (64.3\% [38.8, 83.7]) & $\phantom{-}0.19 \pm 0.29$ & $0.35$ \\
Rater 3, Rater 4 &  26 & 17 (65.4\% [46.2, 80.6]) & $\phantom{-}0.12 \pm 0.24$ & $0.38$ \\
Rater 3, Rater 5 &  13 & 12 (92.3\% [66.7, 98.6]) & $\phantom{-}0.63 \pm 0.36$ & $0.30$ \\
Rater 4, Rater 5 &  13 &  9 (69.2\% [42.4, 87.3]) & $\phantom{-}0.40 \pm 0.25$ & $0.18$ \\
\bottomrule
\end{tabular}
\end{table}

\paragraph{Choice breakdown on overlapping triplets.} Most disagreements involve one rater choosing the CIMP GAN where the other chose Noise2Void, rather than the symmetric reverse. The full $2{\times}2$ counts per pair are in Table~\ref{tab:user_study_pairs}, where columns are the second rater's choice and rows are the first rater's choice.

\begin{table}[h]
\centering
\caption{Per-pair $2{\times}2$ counts of chosen denoiser on overlapping triplets. Rows are the first rater's choice, columns are the second rater's choice.}
\label{tab:user_study_pairs}
\footnotesize
\begin{tabular}{llrr}
\toprule
Pair & First $\backslash$ Second & CIMP & N2V \\
\midrule
Rater 1, Rater 2 & CIMP & 60 &  7 \\
                 & N2V  & 17 & 17 \\
\midrule
Rater 1, Rater 3 & CIMP & 52 &  4 \\
                 & N2V  & 13 &  7 \\
\midrule
Rater 1, Rater 4 & CIMP & 27 & 24 \\
                 & N2V  & 10 & 12 \\
\midrule
Rater 1, Rater 5 & CIMP & 16 &  5 \\
                 & N2V  &  9 &  6 \\
\midrule
Rater 2, Rater 3 & CIMP & 28 &  2 \\
                 & N2V  &  6 &  4 \\
\midrule
Rater 2, Rater 4 & CIMP & 16 & 11 \\
                 & N2V  &  4 &  2 \\
\midrule
Rater 2, Rater 5 & CIMP &  7 &  2 \\
                 & N2V  &  3 &  2 \\
\midrule
Rater 3, Rater 4 & CIMP & 15 &  7 \\
                 & N2V  &  2 &  2 \\
\midrule
Rater 3, Rater 5 & CIMP & 11 &  1 \\
                 & N2V  &  0 &  1 \\
\midrule
Rater 4, Rater 5 & CIMP &  5 &  1 \\
                 & N2V  &  3 &  4 \\
\bottomrule
\end{tabular}
\end{table}

\paragraph{Unanimity at $\geq 3$ raters.} Of the $92$ triplets judged by at least three raters, $47$ ($51.1\%$, Wilson $95\%$ CI $[41.0, 61.1]$) showed unanimous agreement among the raters who saw them, with the remaining $45$ split. A two-sided binomial test against the $50\%$ null rate is not significant ($p = 0.92$), so unanimous agreement is no more or less common than a coin flip on the multiply-judged subset.

Four of the five raters individually prefer the CIMP GAN at $68$--$86\%$ while one rater is indistinguishable from chance, and only the Rater 1 vs.\ Rater 2 pair has a $\kappa$ that is significantly different from zero at $p < 0.05$. Other $\kappa$ values are positive but small with confidence intervals straddling zero, so the per-triplet ratings are not interchangeable.

\section{Limitations}
\label{app:limitations}

\textbf{Boundary conditions are missing from the training data.} Operators do not routinely record images with clipped intensities, saturated detectors, or other extreme boundary conditions, because such acquisitions are considered bad. As a result, the upper and lower bounds of each acquisition parameter are under-represented in the dataset, and the model has no examples from which to learn the corresponding image behaviours. This is visible in the gain sweep (Figure~\ref{fig:exp_eval_gain}), where the model correctly reproduces the trend of increasing brightness and noise but fails to saturate at the high end, and in the offset sweep (Figure~\ref{fig:exp_eval_offset}), where extreme values are poorly reproduced.

\textbf{Scale and diversity relative to computer-vision datasets.} To our knowledge, the 7{,}330 images released here constitute the largest paired image--metadata HAADF-STEM dataset published to date, and far exceed any comparable precedent in materials electron microscopy. It is nonetheless modest and narrow by computer-vision standards: ImageNet~\citep{deng2009imagenet} contains 1.28 million natural images and has supported three orders of magnitude of model scaling beyond what our corpus can support. Future work will include amassing a higher volume and more diverse corpus of data from further research facilities.

\section{Compute Resources}
\label{app:compute}

All experiments ran on a SLURM-managed cluster of NVIDIA H100 80\,GB GPUs (one p5.48xlarge node per job, 8 GPUs available per node).

\paragraph{CIMP encoder training.} The adopted ResNet-18 configuration (crop size 256, metadata encoder $256\times 3$, embedding dimension 128, effective batch size 512, 1000 epochs) used 8 H100s and took roughly 12 hours of wall-clock time. The ViT-pretrained variant at the same crop size and batch size used the same hardware and ran for a similar duration. Smaller ablation runs reported in Appendix~\ref{app:pretraining_ablations} (backbone, crop size, metadata-encoder capacity, batch size sweeps) used 1 to 8 H100s each and ran between 2 and 16 hours depending on configuration.

\paragraph{Style-transfer GAN training.} The adopted configuration (StyleUNet with FiLM conditioning, base filters 32, crop 256, batch 32, 250 epochs, $\lambda_{\text{cyc}}=0.3$) used 1 NVIDIA RTX PRO 4000 Blackwell and took roughly 11 hours. Each of the 11 ablation runs in Appendix~\ref{app:gan_ablations} ran for 100 epochs on 1 NVIDIA RTX PRO 4000 Blackwell, taking roughly 5 hours each.

\paragraph{Linear probing and similarity evaluation.} Embedding extraction over the full $7{,}330$-image corpus and Ridge regression fitting take under 5 minutes on 1 H100. The bootstrap uncertainty estimates reported in Table~\ref{tab:linear_probe} use 2000 resamples of the validation predictions and add roughly 1 minute on CPU.

\paragraph{Denoising baselines.} Noise2Void retraining on our train split (12 epochs, batch 32, $256\times 256$ patches) took 20 minutes on 1 H100. Noise2Atom retraining on our train split ($256\times 256$ patches, 1024 cycle-WGAN loops, batch 3) took roughly 18 hours on 1 H100; an earlier 128$\times$128 retraining run took 10 hours. The three-way comparison (CIMP GAN, Noise2Void, Noise2Atom inference plus radial-spectrum analysis) over 100 held-out frames takes 7 minutes on 1 H100.

\paragraph{Total project compute.} The experiments reported in this paper consumed roughly 800 GPU-hours on H100s. Failed and exploratory runs that did not make it into the paper (loss-weight searches, alternative backbones, earlier dataset versions) account for roughly another 1500 GPU-hours.

\newpage

\end{document}